%% file: root.tex
\documentclass[letterpaper, 10 pt, conference]{ieeeconf}  

\IEEEoverridecommandlockouts                              

\usepackage{epsfig} 
\usepackage{graphicx} 
\usepackage{times} 
\usepackage{amsmath} 
\usepackage{amssymb}  
\usepackage{cite}

\input{custom/packages.tex}
\input{custom/commands.tex}

\usepackage[pagebackref=true,breaklinks=true,colorlinks,bookmarks=false]{hyperref}

\title{\LARGE \bf \approachname: A Complete Pipeline for Dense Correspondence-based\\6D Object Pose Estimation without CAD Models}

\author{Francesco Milano$^{1}$, Jen Jen Chung$^{2}$, Hermann Blum$^{1}$, Roland Siegwart$^{1}$, Lionel Ott$^{1}$
\thanks{$^{1}$ETH Zurich, Switzerland, $^{2}$The University of Queensland, Australia.
}
\thanks{\noindent
This work has
received funding from
the European Union's Horizon 2020 research and innovation program under grant agreement No. 101017008.
}
}

\begin{document}

\begin{textblock}{13}(1.5,0.25)
\centering \noindent\footnotesize © 2024 IEEE.  Personal use of this material is permitted.  Permission from IEEE must be obtained for all other uses, in any current or future media, including reprinting/republishing this material for advertising or promotional purposes, creating new collective works, for resale or redistribution to servers or lists, or reuse of any copyrighted component of this work in other works.
\end{textblock}
\maketitle
\thispagestyle{empty}
\pagestyle{empty}

\input{sections/1_abstract.tex}

\input{sections/2_introduction.tex}
\input{sections/3_related_work.tex}
\input{sections/4_method.tex}
\input{sections/5_experiments_camera_ready.tex}
\input{sections/6_discussion_and_conclusion.tex}





{\small
\bibliographystyle{IEEEtran}
\bibliography{egbib_no_links}
}

\end{document}

%% file: custom/packages.tex
\usepackage{pifont}%
\usepackage{booktabs}

\usepackage{soul}

\usepackage{pdflscape}

\usepackage{multirow,makecell}
\usepackage{array}
\newcolumntype{S}{>{\centering\arraybackslash}m{1.5cm}}
\newcolumntype{L}{>{\centering\arraybackslash}m{5cm}}

\usepackage{colortbl}
\usepackage[x11names, table]{xcolor}

\usepackage[referable]{threeparttablex}

\usepackage{siunitx}

\usepackage{xspace}

\usepackage{lipsum}

\usepackage{graphicx}
\usepackage{tabularx}

\usepackage{svg}

\usepackage{rotating}

\usepackage{moresize}

\usepackage[caption=false]{subfig}

\usepackage[absolute]{textpos}

%% file: custom/commands.tex
\newcommand{\approachname}{NeuSurfEmb\xspace}

\makeatletter
\DeclareRobustCommand\onedot{\futurelet\@let@token\@onedot}
\def\@onedot{\ifx\@let@token.\else.\null\fi\xspace}

\def\eg{\emph{e.g}\onedot} 
\def\ie{\emph{i.e}\onedot} 
 
\def\etc{\emph{etc}\onedot}

\makeatother

\newcommand\RotText[1]{\rotatebox{90}{\parbox{1.5cm}
{\centering#1}}}

%% file: sections/1_abstract.tex
\begin{abstract}
State-of-the-art approaches for 6D object pose estimation assume
the availability of
CAD models and require
the user to
manually
set up
physically-based rendering (PBR) pipelines
for synthetic training data generation.
Both
factors limit the application of these methods in real-world scenarios. In this work, we
present a
pipeline
that does not require CAD models and
allows
training a state-of-the-art pose estimator
requiring only a small set of real 
images as input.
Our method is based on a NeuS2~\cite{Wang2023NeuS2} object
representation,
that we learn through a semi-automated 
procedure
based on Structure-from-Motion (SfM) and object-agnostic segmentation.
We exploit the novel-view synthesis ability of NeuS2 and simple
\textit{cut-and-paste} augmentation to automatically generate photorealistic object renderings, which we
use to train the 
correspondence-based
SurfEmb~\cite{Haugaard2022SurfEmb} pose estimator.
We evaluate our
method
on the LINEMOD-Occlusion dataset, extensively
studying
the impact of its
individual
components
and showing
competitive performance with respect to
approaches
based on
CAD models and PBR data.
We additionally demonstrate the ease
of use and effectiveness
of our pipeline
on self-collected real-world objects,
showing
that our
method
outperforms state-of-the-art
CAD-model-free approaches,
with
better accuracy and robustness to mild occlusions.
To allow the robotics community to benefit from this system,
we will publicly release it
at \url{https://www.github.com/ethz-asl/neusurfemb}.
\end{abstract}

%% file: sections/2_introduction.tex
\section{Introduction}
Estimating the 6D pose of objects from image observations is a long-standing problem in computer vision 
and of broad interest to several important real-world applications,
including robotic manipulation~\cite{Florence2018DONs, Manuelli2019kPAM, Tremblay2018DOPE}, augmented reality~\cite{Hagbi2010ShapeRecognitionPoseEstimationMobileAugmentedReality, Marchand2015PoseEstimationAugmentedRealitySurvey, Su2019DeepMultistateAugmentedRealityAssembly}, and object-level mapping~\cite{Ruenz2017Co-Fusion, SalasMoreno2013SLAM++}.

Many of the current state-of-the-art approaches
require a high-fidelity, often textured, CAD model of an object to estimate its pose~\cite{
Haugaard2022SurfEmb, Labbe2020CosyPose, Li2019CDPN, Shugorov2022OSOP,
Zakharov2019DPOD}. While the datasets used in
recent evaluation benchmarks~\cite{Hinterstoisser2012LineMod, Hodan2017TLESS, Hodan2018BOPBenchmark} provide this information, in practical real-world applications, obtaining an accurate, textured CAD reconstruction is often non-trivial, usually requiring manual design or specialized equipment for data collection
or
intensive
post-processing~\cite{Hodan2017TLESS, Rennie2016RU-APCDataset}.
Moreover,
the vast majority of the proposed
approaches are trained on large synthetic datasets 
generated
through
physically-based rendering (PBR)
pipelines~\cite{Hodan2020BOPChallenge, Hodan2019PhotorealisticImageSynthesisObjectInstanceDetection}.
These pipelines
produce
photorealistic
images
with physically accurate modelling of
light
and material properties; 
however, they
require
textured CAD models and
proper
setup and
parameter configuration
from
an experienced user,
which
makes
their
application to
a
new,
real-world object
non-straightforward.

To be of
practical use 
for a real-world system, \eg, a
robot or
an augmented
reality headset,
a pose estimation algorithm 
would
instead
ideally
require the user to provide
only
a small set of observations of
an
object of
interest.
With this goal in mind,
a number of
\emph{model-free}
approaches for 6D pose estimation have
been proposed,
which
typically
construct
a
Structure-from-Motion (SfM)-based model of the object and 
later relocalize
the camera with respect to
it~\cite{Liu2022Gen6D, He2022OnePose++}.
While
these methods
allow relaxing the assumption of a CAD model,
they
tend to be less accurate than
state-of-the-art methods leveraging CAD models and PBR data,
and
typically show limited robustness to occlusions.

In this work, we 
propose a framework that allows training a
pose estimator for
real-world objects
without requiring a CAD model or
a PBR
synthetic dataset,
while still
achieving
performance
comparable
to state-of-the-art
approaches that require the latter.
Our method is provided in the form of a semi-automated pipeline that
simply
requires
a sparse set of image observations of the object and a
bounding box
indicating the object of interest in
one frame.
After training, our system
allows estimating the 6D pose of the object from a single RGB image, with optional depth-based refinement.
We 
use a
neural implicit surface reconstruction method (NeuS2~\cite{Wang2023NeuS2})
as the underlying representation for the object of
interest.
Using SfM in combination with state-of-the-art object-agnostic segmentation~\cite{Kirillov2023SAM} and tracking~\cite{Cui2022MixFormer}, we automatically estimate poses and object masks for each of the
reference
images.
We
then use these frames to train
an object-level NeuS2,
which
compactly and
accurately
reconstructs
the object, effectively replacing
a
CAD 
model.
At the same time,
we show that
NeuS2 can replace
a more
involved
PBR
pipeline
and efficiently generate renderings to train
a pose estimator.
For the latter,
we leverage SurfEmb~\cite{Haugaard2022SurfEmb}, a recent 
method
based on dense correspondences.

We evaluate our approach on the
LINEMOD-Occlusion~\cite{Brachmann2014LineModOcclusion} dataset, and conduct extensive ablations
to highlight the effect
of each component in our pipeline. We additionally demonstrate our
method
on a set of real-world objects,
performing
both qualitative and quantitative evaluations against state-of-the-art
baselines
that,
like our method,
are applicable
when no CAD models are available.
Our
pipeline,
which we name \approachname, achieves comparable performance to
CAD-model-based methods
and outperforms
previous
CAD-model-free
approaches.

In summary,
our main contributions
are the following:
\begin{enumerate}
    \item[i.] A
    pipeline for 6D object pose estimation
    requiring only a small set of real RGB images as input.
    \item[ii.] Extensive ablation studies on our object representation, training data, and other components of our pipeline.
    \item[iii.] Evaluation on 
    a standard dataset
    and on
    real-world data, showing that our approach achieves competitive performance against state-of-the-art methods, while being applicable to real-world objects.
    \item[iv.] An open-source
    implementation 
    to easily train and deploy our
    pipeline
    for novel objects.
\end{enumerate}

%% file: sections/3_related_work.tex
\section{Related work}
\subsection{CAD model-based object pose estimation}
A large number of state-of-the-art approaches for 6D object pose estimation rely on the assumption that a
CAD model of the object of interest is available.
On one side, the CAD model is used to generate synthetic data
through photorealistic rendering pipelines, often based on PBR~\cite{Denninger2023BlenderProc2, Hodan2019PhotorealisticImageSynthesisObjectInstanceDetection}.
For this reason, high-fidelity object texture is needed to produce high-quality renderings with limited domain gap with respect to the real data on which the trained algorithms are evaluated.
On the other side, the CAD model is used during the training of the pose estimation algorithm. For instance,
keypoint-based methods~\cite{Rad2017BB8,
Tekin2018YOLO6D,
Peng2019PVNet} predict the 2D location of pre-defined salient points, and use the object model to estimate the object pose
based on
2D-3D 
correspondences.
Coordinate-based methods~\cite{Li2019CDPN, Zakharov2019DPOD, Wang2019NOCS} predict 
a
3D coordinate, defined
according to
the object model,
for
each pixel in the input image.
\cite{Shugorov2022OSOP} and~\cite{Labbe2022MegaPose} render reference images from the textured CAD 
model,
use pre-trained networks
to find the
reference image
that best matches the test
one,
and subsequently refine the
pose.
Correspondence-based methods~\cite{Haugaard2022SurfEmb, Huang2022NCF},
which  achieve state-of-the-art robustness to occlusions,
compute \emph{dense} correspondences between
image
pixels and 3D points 
on the object 
model.
We base our pose estimation algorithm on SurfEmb~\cite{Haugaard2022SurfEmb}, a
recent method from the latter category, and relax its assumptions of a
CAD model and PBR synthetic dataset.

\subsection{CAD model-free object pose estimation}
SfM-based methods are the current state of the art for CAD model-free object pose estimation~\cite{He2022OnePose++, Liu2022Gen6D}.
These methods assume
a set of reference images, which are used to construct a sparse~\cite{Liu2022Gen6D} or semi-dense~\cite{He2022OnePose++} point cloud, using SfM. Together with the reference images, the point cloud acts as a 3D model,
and
is used to estimate the pose of the object in a test image
through
feature matching
and PnP.
SfM-based methods
tend to be less accurate and robust to occlusions than state-of-the-art, CAD-model-based methods, but
can easily be applied
in a real-world scenario where no CAD models are available.
We follow a similar setup
in our method, assuming a given set of reference images and running SfM on them;
however,
we base
our object model on
NeuS2 rather than a point cloud.

\subsection{Object pose estimation via neural implicit representations}
Similarly to our method, the
recent
BundleSDF~\cite{Wen2023BundleSDF} 
and
TexPose~\cite{Chen2023TexPose}
perform
6D pose estimation based
on a neural implicit object representation. However,
BundleSDF
additionally
requires depth
images
as input,
while
TexPose
assumes
a CAD model and a PBR synthetic dataset.

\input{figures/method_overview}

%% file: figures/method_overview.tex
\begin{figure*}[th!]
    \centering
    \includegraphics[width=0.95\linewidth]{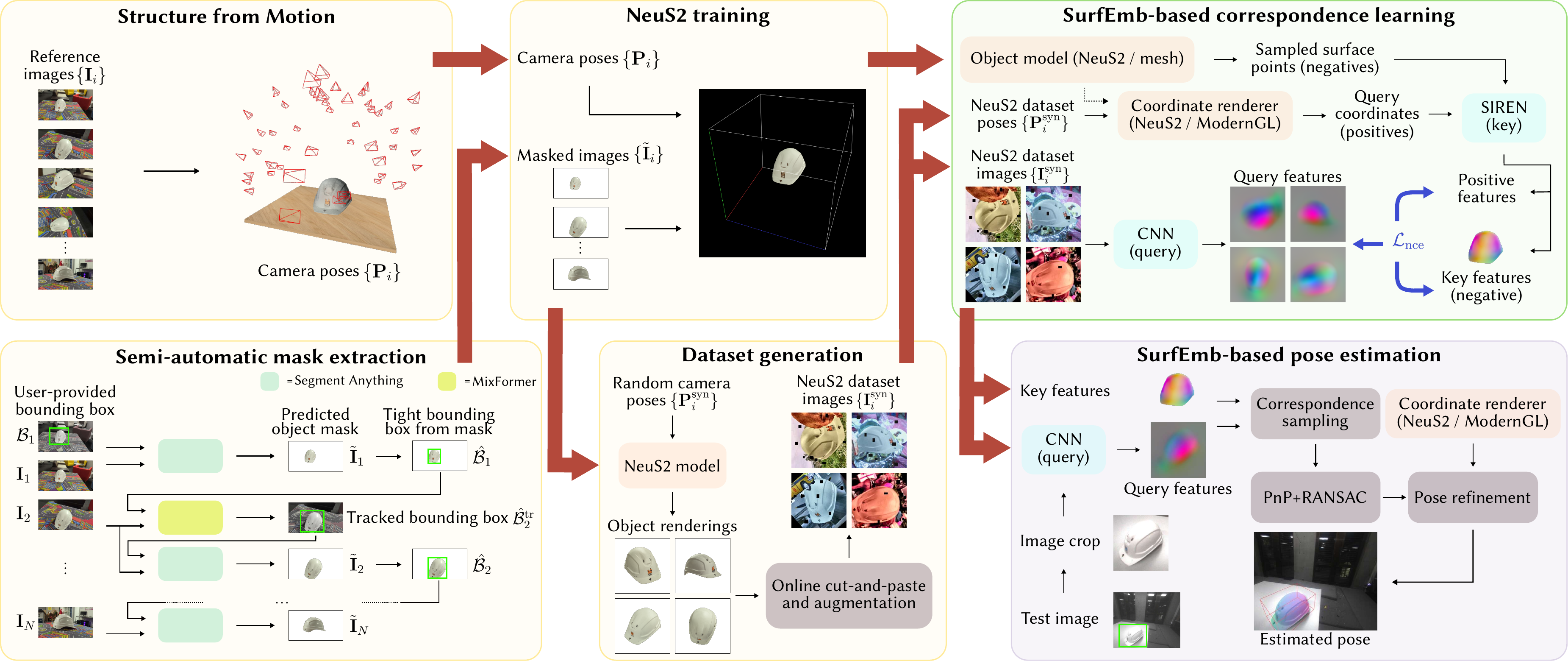}
    \caption{Overview of the proposed method. Starting from a set of reference images $\{\mathbf{I}_i\}$ around the object of interest, and using Structure-from-Motion and a pipeline based on
    Segment Anything~\cite{Kirillov2023SAM} and the object tracker MixFormer~\cite{Cui2022MixFormer} to estimate corresponding camera poses $\{\mathbf{P}_i\}$ and object masks $\{\tilde{\mathbf{I}}_i\}$, we construct an object model and synthesized dataset by training a NeuS2~\cite{Wang2023NeuS2} model and generating renderings from novel views $\{\mathbf{P}^\textrm{syn}_i\}$ (yellow boxes). We use the generated object model and synthesized dataset,
    augmented
    online using cut-and-paste~\cite{Dwibedi2017CutPasteLearn} to simulate occlusions and background variations, to learn feature-based dense 2D-3D correspondences based on SurfEmb~\cite{Haugaard2022SurfEmb} (green box). We then estimate the object pose in a test image by sampling correspondences based on the learned object features and the predicted image features and using PnP with RANSAC and pose refinement
    (purple
    box).}
    \label{fig:method_overview}
    \vspace{-15pt}
\end{figure*}

%% file: sections/4_method.tex
\section{Method\label{sec:method}}
An overview of our method is shown in Fig.~\ref{fig:method_overview}. We first present the steps to generate an object model and a synthesized dataset based on NeuS2 (yellow boxes), then we detail the steps to learn 2D-3D correspondences (green box), and finally we describe the procedure to estimate the pose of the object of interest in a test image
(purple
box).

\subsection{NeuS2-based object model and dataset\label{sec:model_and_dataset_building}}
Similarly to other CAD-model-free methods~\cite{Liu2022Gen6D, He2022OnePose++}, in our setting we assume
to have available
a small set of $N$ images $\{\mathbf{I}_i\}$
(with $N\approx100$),
captured at roughly uniformly-distributed viewpoints around the object. 
These
images
are
used to
construct
a reference object model with respect to which the object pose in test images is 
later
estimated.
We perform COLMAP~\cite{Schoenberger2016COLMAP}-based SfM to
retrieve
camera poses $\{\mathbf{P}_i\}$ associated to the reference images $\{\mathbf{I}_i\}$.
However, in contrast to the sparse or semi-dense cloud of triangulated points that form the SfM-based object models in~\cite{Liu2022Gen6D, He2022OnePose++}, 
we
learn a dense object model based on NeuS2~\cite{Wang2023NeuS2}.

To this
end,
we first extract object masks from the reference images using a semi-automatic pipeline based on 
Segment Anything (SAM)~\cite{Kirillov2023SAM} and 
MixFormer~\cite{Cui2022MixFormer}. We assume that the object of interest is not occluded in the reference images, that the latter are extracted from a temporal sequence, and that
a bounding box $\mathcal{B}_1$ around the object in the first frame
$\mathbf{I}_1$ is provided.
$\mathcal{B}_1$ is used to prompt SAM for an object mask for $\mathbf{I}_1$. We then fit a tight bounding box $\hat{\mathcal{B}}_1$ around the predicted object mask $\tilde{\mathbf{I}}_1$, and provide it to MixFormer as initialization for the subsequent frame $\textbf{I}_2$. The bounding box $\hat{\mathcal{B}}^\textrm{tr}_2$ returned as output by MixFormer is then used as prompt for SAM to extract a mask $\tilde{\mathbf{I}}_2$ for the image $\mathbf{I}_2$, and the process is repeated for all
the frames $\mathbf{I}_t$, with
$t\in\{2,\dots, N\}$, using $\hat{\mathcal{B}}_{t-1}$ as initialization for MixFormer and the tracked bounding box $\hat{\mathcal{B}}^\textrm{tr}_t$ as a prompt for SAM.
We find that this process accurately segments the object
in the majority of the frames,
and
we provide additional tools based on SAM to refine
the few
poorly segmented frames, for instance by allowing multiple prompts per frame.

We use the extracted masked images $\{\tilde{\mathbf{I}}_i\}$ and the estimated camera poses $\{\mathbf{P}_i\}$ to train an object-level NeuS2 model through inverse volume rendering, using its standard supervision setting that combines a robust color loss 
with
a regularization Eikonal loss~\cite{Wang2023NeuS2}.
By relying on an underlying signed distance field (SDF) and employing an unbiased volume rendering formulation~\cite{Wang2021NeuS}, NeuS2 is able to accurately reconstruct the object surface, which we extract either as a mesh model, using Marching Cubes~\cite{Lorenson1987MarchingCubes}, or as a point cloud, by rendering per-pixel 3D coordinates from different viewpoints and aggregating them.

At the same time, we exploit the ability of NeuS2 to synthesize novel high-fidelity views of the object to efficiently produce renderings $\{\mathbf{I}^\textrm{syn}_i\}$ of the object from camera poses $\{\mathbf{P}^\textrm{syn}_i\}$ that we randomly sample from the top hemisphere around the object.
To ensure that the coordinate frame of the NeuS2 model, and consequently the rendered images, align with the natural 
frame of the object, we perform NeuS2 training in two steps: 
(i) We re-orient the reference SfM camera poses $\{\mathbf{P}_i\}$ based on their viewing directions so that the NeuS2 coordinate frame roughly coincides with the center of the observed object;
(ii) We extract a point cloud from the NeuS2 model trained in the first step and fit a 3D oriented bounding box to it.
We redefine the NeuS2 coordinate frame to be centered in the bounding box and aligned with its axes. We re-align the reference camera poses accordingly, and then re-train the NeuS2 model with the new coordinate frame.
Since the subsequent training steps require the images to be cropped
at a fixed square resolution
around the object,
to maintain efficiency and image quality
we directly render the synthesized images $\{\mathbf{I}^\textrm{syn}_i\}$ cropped around the object.
For each viewpoint $\mathbf{P}^\textrm{syn}_i$, we
do so by reprojecting the object point cloud onto the image plane, fitting a 2D bounding box around it and adapting the camera intrinsics to render only within the bounding box.

\subsection{Correspondence training}
We use the 
NeuS2-based object model and
dataset to learn 2D-3D correspondences 
using
SurfEmb~\cite{Haugaard2022SurfEmb}.
In particular, we instantiate a convolutional neural network (CNN) -- also referred to as \emph{query network} -- and a coordinate field based on SIREN~\cite{Sitzmann2020SIREN} -- \emph{key network}, to respectively return a high-dimensional feature $\mathbf{f}_q(\mathbf{p})\in\mathbb{R}^d$ for each pixel $\mathbf{p}$ in the input image and a feature vector $\mathbf{f}_k(\mathbf{x})\in\mathbb{R}^d$ for each 3D point $\mathbf{x}$ on the object surface. For each input image, we additionally render
3D coordinates
corresponding to each
pixel,
either by using
a
mesh-based
ModernGL~\cite{Dombi2020ModernGL} 
renderer (as in SurfEmb), that we provide with the Marching Cubes mesh extracted from NeuS2,
or
by
directly rendering the coordinates using NeuS2. We then query the key network at the rendered 3D coordinates and train the two networks using
a contrastive Info-NCE loss $\mathcal{L}_\textrm{nce}$~\cite{VanDenOord2018InfoNCE}.
Like
SurfEmb, we additionally output from the query network an object mask, supervised with a cross-entropy loss with respect to the ground-truth object
mask.

Crucially,
since the rendered images
$\{\mathbf{I}^\textrm{syn}_i\}$ 
only contain the object of interest, we apply
online \textit{cut-and-paste}~\cite{Dwibedi2017CutPasteLearn} augmentation to the foreground and background of each training image, to simulate occlusions and 
background variations, respectively.
For the background, we
use,
with equal probability, random noise and a crop of an image sampled from the PASCAL-VOC dataset~\cite{Everingham2012PASCALVOC}; for the foreground, we randomly select an instance from a
PASCAL-VOC image and
place it
on the rendered image so that the percentage of occluded object pixels varies
between $20\%$ and $70\%$.
We apply extensive color augmentation and in-plane affine transformations, as done in SurfEmb, and additionally employ white-balancing augmentation using the method of~\cite{Afifi2019WBAugmenter},
which we find beneficial
for
the 
generalization of the method.
\subsection{Pose estimation}
Given a test image containing the object of interest, we estimate the 6D pose of the object with respect to the camera using 
correspondence-based method 
of~\cite{Haugaard2022SurfEmb}. In particular, assuming a 2D bounding box provided by an external object 
detector,
we crop and rescale the input image to the
resolution used during training and feed it to the query network. We then
compute the similarity between the output query features, weighted by the predicted object mask, and the features returned by the key network for a uniform set of surface points. The resulting similarity matrix is used to sample 2D-3D correspondences using importance sampling. A set of candidate poses is obtained from these correspondences using PnP+RANSAC, and a refinement step is applied to the best-scoring pose using a coordinate renderer (we refer the reader to Sec.~3.3 of~\cite{Haugaard2022SurfEmb} for exact details). Similarly to the correspondence learning step, in our setup
either a mesh-based renderer or NeuS2 can be used as coordinate renderer.
Finally, an additional refinement step
can be performed if a depth image is available.

%% file: sections/5_experiments_camera_ready.tex
\section{Experiments and Results}
\label{sec:experiments}
In the
following,
we assess our method's performance and the impact of its components. We cover evaluation metrics and training details in
Sec.~\ref{sec:experimental_setup}.
We analyze the reconstruction quality of NeuS2 in Sec.~\ref{sec:reconstruction_quality}, our pose estimation performance on LINEMOD-Occlusion in Sec.~\ref{sec:linemod_occlusion_experiments}, and the impact of our NeuS2-based object model and dataset in Sec.~\ref{sec:ablation_object_model_and_data_generation}. Lastly, in Sec.~\ref{sec:real_world_experiments}, we test our method on self-collected real-world data.
\subsection{Experimental setup}~\label{sec:experimental_setup}
For the experiments on LINEMOD-Occlusion, we report the standard BOP Average Recall 
error measure,
$\mathrm{AR}_\mathrm{BOP}$,
which
takes into account both object symmetries and occlusions in the scene~\cite{Hodan2020BOPChallenge}. For the real-world experiments, we use the standard 
$\mathrm{ADD(-S)}$~\cite{Hinterstoisser2012LineMod, Hodan2016OnEvaluation6DObjectPoseEstimation}
and
$\SI{5}{cm},\ \SI{5}{{}^\circ}$~\cite{Shotton2013SceneCoordinateRegressionForests}
metrics for the scenes with no occlusions and $\mathrm{AR}_\mathrm{BOP}$ and $\SI{5}{cm},\ \SI{5}{{}^\circ}$ for the scenes with occlusions.

We train NeuS2 for $\num{20000}$ steps.
For each object, we generate $\num{10000}$
images at
a 
resolution of $224\times224$ pixels.
For correspondence learning, we use the same
network architectures as
in SurfEmb,
and
train
a model for each object
for $50$ epochs, to achieve a similar number of
iterations as in the original method~\cite{Haugaard2022SurfEmb}.
NeuS2 training takes $\SI{10}{\minute}$, dataset generation
$20$
to $\SI{30}{\minute}$, and correspondence learning
$1$
to $1.5$ days,
on a single NVIDIA RTX $2080$ Ti GPU.
\subsection{NeuS2 reconstruction quality~\label{sec:reconstruction_quality}}
Given the correspondence-based nature of our pose estimation method, the geometric accuracy of the 3D object model
might play
an important role in the quality of the estimated pose, since selecting potentially inaccurate 3D surface points as samples for the PnP+RANSAC step
might
directly result
in errors in the predicted pose.
To investigate the impact of the model accuracy,
we first
assess
the reconstruction quality of our NeuS2-based 3D models. We evaluate it for the $8$ objects in the LINEMOD-Occlusion dataset~\cite{Brachmann2014LineModOcclusion} and report
it as the 
forward Chamfer distance between the NeuS2 mesh reconstructions and the corresponding ground-truth CAD models available in the dataset.
For each object, we train a NeuS2 model using the images and ground-truth poses and masks from the corresponding scene in the occlusion-free LINEMOD dataset~\cite{Hinterstoisser2012LineMod}.
As shown in Fig.~\ref{fig:failure_mode_reconstruction_linemod}, NeuS2 achieves very accurate reconstruction (in the order of $\SI{1}{\milli\meter}$ or
lower
reconstruction error) for $5$ out of $8$ objects.
For
the remaining $3$ objects,
a closer look at the reconstructed surfaces shows that two main failure modes can be highlighted (bottom row): 1. Partial holes inside the
objects tend to not be
properly captured ($\mathrm{can}$, $\mathrm{holepuncher}$); 2. Surface parts that are
not visible
in the training views cannot be reconstructed, as for instance the cone-like structures at the bottom of the $\mathrm{eggbox}$.
We investigate the impact of the reconstruction quality on the pose predictions in
Sec.~\ref{sec:ablation_object_model_and_data_generation}.
\input{figures/failure_mode_reconstruction_linemod}

\subsection{LINEMOD-Occlusion experiments\label{sec:linemod_occlusion_experiments}}
\input{tables/linemod_occlusion}
\input{tables/ablation_object_model_for_pose_estimation}

\input{tables/linemod}
We evaluate the pose estimation performance of
our method
on the LINEMOD-Occlusion dataset, which contains
pose annotations for
a subset of $8$ objects from the original LINEMOD dataset and
in which,
unlike LINEMOD, the objects of interest
present occlusions.
For each object, we train a NeuS2 model following the same setup as in Sec.~\ref{sec:reconstruction_quality}.
Following the standard practice
in the literature~\cite{Haugaard2022SurfEmb},
we use the object detections from CosyPose~\cite{Labbe2020CosyPose} to crop the
test
images for pose estimation.
We compare the performance of our method to state-of-the-art approaches which, unlike our method, assume a CAD model and a 
PBR synthetic dataset.

Table~\ref{tab:linemod_occlusion} shows the results of the evaluation. 
Both
with RGB-only and with RGB-D inputs,
\approachname
achieves comparable performance to several CAD-model-based baselines~\cite{Hodan2020EPOS, Peng2019PVNet} and is
outperformed
by a small margin
by others~\cite{Haugaard2022SurfEmb, Li2019CDPN, Labbe2020CosyPose}.
While the coordinate renderer (cf.
rows $1$ and $2$) 
and the object model
(cf.
rows $2$ and $4$)
both have minimal impact on the output performance, which confirms that NeuS2 is able to
accurately approximate the ground-truth object geometry,
we find that the major differentiating factor 
is the type of
images used
for correspondence learning.
When using PBR images instead of NeuS2-generated ones, our method achieves virtually the same performance as the leading method, SurfEmb (cf.
rows $3$ and $5$).
We hypothesize that
this
performance discrepancy
is largely due to the
way our image generation approach
simulates
occlusions, which appear less realistic compared to those in PBR images.
We validate this hypothesis and further investigate the effect of both
object model and
synthesized images on the final 
performance in the next Section.

\subsection{\hbox{Ablation:\hspace{2pt}Effect\hspace{2pt}of\hspace{2pt}NeuS2-based\hspace{2pt}object\hspace{2pt}model\hspace{2pt}and\hspace{2pt}images}~\label{sec:ablation_object_model_and_data_generation}}
In this ablation study, we
report
the performance for each object individually, to
better capture potential object-specific factors.
Table~\ref{tab:ablation_object_model_and_training_images}
shows
the results of the evaluation, where in the top $4$ rows we evaluate models trained with NeuS2 object models and in the bottom $4$ rows those trained with CAD models.
From the pairs of rows in Tab.~\ref{tab:ablation_object_model_and_training_images}, \ie, $1$-$2$, $3$-$4$, \etc
we can see 
that for all the objects except $\mathrm{eggbox}$, $\mathrm{holepuncher}$, and to some extent $\mathrm{can}$, using a different model for training and pose estimation 
has minimal effect on
the pose estimation performance. This aligns with the results discussed in Sec.~\ref{sec:reconstruction_quality} on the per-object reconstruction quality. 
In addition, it should be noted that similar performance is obtained across all objects when training and pose estimation both use the same model (Neus2 or CAD, \eg, rows $1$ and $5$).
We
attribute this result to the fact that
the key network learns to assign low-norm features to
parts of the object that are
never or only rarely
observed
during
training, which as noted in Sec.~\ref{sec:reconstruction_quality} 
constitute the main factor of
geometric
discrepancy
between
the
two
types of
object
model.
As a result, 3D points on these parts
are
sampled only with low probability during pose estimation, yielding limited inconsistencies between the pose estimates for the two types of model. 
A second point that can be
noted
is that for both
NeuS2 and CAD,
the effect of the training images
is largely dependent on the specific object.
We hypothesize that
this variability is due to a combination of differences in the object textures, which may be simulated more accurately by PBR for certain objects, and of the different effectiveness of
how occlusions are simulated in the two types of images.
To further investigate the impact of the simulated occlusions on the results, we
test
the
models trained for the experiments in Sec.~\ref{sec:experiments} also on the occlusion-free LINEMOD dataset, reporting the performance for the $8$ objects shared across both datasets. We use ground-truth bounding boxes to crop the test images.
The results of this ablation,
reported
in Tab.~\ref{tab:linemod},
show
that when no occlusions are present in the test data, NeuS2-generated images perform
on-par or
at times even
better than PBR
images.
This finding
supports our hypothesis that our \textit{cut-and-paste} strategy for occlusion simulation has margin for improvement, but
at the same time
indicates that our NeuS2-synthesized images are 
overall
effective for training
a
pose estimator, in particular when
very high robustness to occlusions is not the main requirement.
Importantly, we stress that while for this particular ablation the domain of the images used to reconstruct the NeuS2 model and that of the test images coincide, 
during training
our pose estimator is provided
exclusively
with \emph{synthesized} images.
Due to our \textit{cut-and-paste} strategy,
background
and foreground
of the synthesized images used
in correspondence learning
are
significantly different from
those
of the test images, which together with
our
extensive data augmentation ensures that no overfitting to the test images can
occur. 
To further validate this point and show that our method achieves
good
generalization, in the real-world experiments presented in the next Section we
change
lighting conditions, background, and camera characteristics between NeuS2 training and pose 
estimation.
\subsection{Real-world experiments~\label{sec:real_world_experiments}}
\input{figures/neus2_reconstruction_real_world_experiments}
\input{figures/qualitative_results}

To demonstrate the effectiveness
and ease of use
of our method for real-world applications,
we collect recordings of $5$ 
different
objects (cf. Fig.~\ref{fig:neus2_reconstruction_real_world_experiments}) and run our pipeline
as described in Sec.~\ref{sec:method}, 
including
obtaining camera poses
and
extracting object masks to train NeuS2.
Note that the
chosen
objects capture different types of challenges, including low
degree
of texture ($\mathrm{helmet}$, $\mathrm{kettle}$), structural symmetries ($\mathrm{bluebox}$, $\mathrm{greybox}$), and complex geometry ($\mathrm{extinguisher}$).
We compare
our method
to two
recent
state-of-the-art approaches, Gen6D~\cite{Liu2022Gen6D} and OnePose++~\cite{He2022OnePose++}, both of which, similarly to our method, do not require a CAD model of the objects of 
interest. We note that CAD-model-free methods are known in the literature to perform worse than CAD-model-based ones~\cite{Sun2022OnePose, Liu2022Gen6D}, and we therefore only reported
baselines
from the latter category
in the dataset evaluations of Sec.~\ref{sec:linemod_occlusion_experiments}.
However, we select \cite{Liu2022Gen6D}
and~\cite{He2022OnePose++}
for comparison in
our real-world experiments
because they represent the best viable option in
a robotic scenario.
As with our method, both Gen6D and OnePose++ also 
require a set of views from a ``model-training" scene
(cf. also Sec.~\ref{sec:model_and_dataset_building}).

For each object, we collect one video recording of the \emph{model-training} scene (Fig.~\ref{fig:neus2_reconstruction_real_world_experiments}), using a 
regular
consumer-grade smartphone, and
multiple \emph{evaluation} scenes using a FLIR Firefly S camera.
For each object,
in one of the evaluation scenes
the object is shown in isolation and in the remaining ones an additional object is present in the scene,
thereby generating occlusions in several viewpoints (Fig.~\ref{fig:qualitative_results}).
Note that
lighting conditions, background, and color characteristics vary significantly between the model-training and the evaluation scenes, which therefore requires the pose estimation algorithms to be
robust to these factors.
\input{tables/results_real_world_experiments_no_occlusions}
\input{tables/results_real_world_experiments_with_occlusions}
To obtain
ground-truth camera-to-object poses, needed for evaluation,
we track the camera
pose
in the evaluation scenes
through a marker-based
motion capture
system (Vicon).
We then convert the camera-to-marker pose to a camera-to-object pose 
by defining an object-centered coordinate frame
for each object,
keeping the position of each object constant across the evaluation recordings, and exploiting the fact that the tracking system coordinate frame stays fixed.
Note that
the coordinate frame of each object in the model-training scene is in general different from the corresponding one in the evaluation scenes, since the model-training frame is
based on the one returned by the SfM pipeline, which is
arbitrarily defined.
Additionally,
given the
inherent
scale ambiguity
of
monocular algorithms like SfM,
the reference model-training poses, and
consequently
the output estimated poses, are not expressed in
meters,
unlike the
poses returned by the Vicon system.
To register the two coordinate 
frames and estimate
the scale conversion factor,
we
train
a
NeuS2 model for both the model-training and the single-object evaluation scene, and
we
estimate the scaled transform between the two coordinate frames 
through
Iterative Closest Point (ICP) registration between the point clouds of the two NeuS2 models.
For both the model-training and the evaluation scenes, we record our videos at 
a
resolution
of
$1920\times\SI{1080}{px}$ and 
sample
the recording to obtain approximately $100$ frames.

We report the results of our evaluation in 
Tables
\ref{tab:results_real_world_experiments_no_occlusion} and~\ref{tab:results_real_world_experiments_with_occlusions}, where we present the performance on each individual object, as well as averaged over all the objects based on their occurrence, for Tab.~\ref{tab:results_real_world_experiments_no_occlusion}.
For the occluded scenes, we only consider an image for evaluation if at least $10\%$ of the object surface is visible.
Both baselines implement
their
own object detector
and
a tracking 
module,
both of which might 
introduce additional failures.
To ensure a fair evaluation focused uniquely on
the pose
estimators,
we
therefore
provide ground-truth bounding boxes
for each frame
to all the methods,
but additionally report the performance of the baselines with their original
setup.
We
compute
the ground-truth bounding boxes in each evaluation image
by rendering object masks using
the NeuS2 models
and the ground-truth camera poses and fitting a
bounding box to the renderings.

Fig.~\ref{fig:qualitative_results} shows qualitative examples
of the estimated poses
(columns $1$-$3$ for the single-object scenes and column $4$-$6$ for
those
with occlusions). Both Gen6D and OnePose++ achieve comparable or even slightly better performance than 
our method
on the two symmetrical and geometrically regular objects ($\mathrm{bluebox}$ and $\mathrm{greybox}$).
However,
their performance drops significantly on the remaining objects, which present either more complex geometry or a lower amount of texture. We note in particular that Gen6D
achieves low performance on the texture-poor $\mathrm{helmet}$
and
fails 
almost
completely for the $\mathrm{extinguisher}$ (see Tab.~\ref{tab:results_real_world_experiments_no_occlusion}). OnePose++
achieves slightly better performance
on
$\mathrm{helmet}$, but
fails to make use of the
low-texture
handle in $\mathrm{kettle}$
to discriminate between similar viewpoints,
thereby returning poses with rotational errors. 
Overall,
\approachname
outperforms both Gen6D and OnePose++
by a significant margin across the objects.

Since as observed in the literature~\cite{Hodan2018BOPBenchmark}, the \hbox{$\mathrm{ADD}$-$\mathrm{S}$} metric tends to return 
large values for symmetric objects and might therefore not be indicative enough of the performance, we
additionally report the recall of 
$\SI{5}{\centi\meter},\ \SI{5}{{}^\circ}$,
thereby not taking symmetries into account. We find that under this metric,
\approachname
achieves better accuracy than the baselines for $\mathrm{bluebox}$, that is, it returns a prediction from the correct side of the object more often than from the opposite one. This indicates that a denser model and explicit image-based training
can use texture information to disambiguate object views that appear identical when only considering geometry.
Nonetheless,
while still achieving close-to-optimal accuracy,
our method is slightly outperformed by the baselines for $\mathrm{greybox}$ also under the $\SI{5}{\centi\meter},\ \SI{5}{{}^\circ}$ metric;
we
notice
that this
performance gap stems mostly from a
systematic
rotational error that our method produces from specific viewpoints.

A relevant
observation, reflected both in the
quantitative and qualitative results,
is that when enabling tracking, as in their original setup, both baselines
generally
achieve lower performance. In particular, Gen6D tends to mistakenly track the foreground object in place of the occluded one.  
An additional element that we notice is that even when providing ground-truth bounding boxes, OnePose++ internally recrops the object detections, which often causes
important object features to be ruled out from triangulation,
particularly for tall objects such as the $\mathrm{extinguisher}$. We note that simply adapting
their
code to keep the full object visible when recropping largely increases the performance on these objects, 
although the accuracy remains
lower than
that of
our method.

Finally, we
observe
that both Gen6D and OnePose++ have limited ability to handle occlusions, returning significantly inaccurate poses for the occluded object also under relatively mild occlusions (see columns $4$ and $6$ in Fig.~\ref{fig:qualitative_results}). In contrast,
\approachname
shows greater robustness to occlusions, which is also reflected in the better quantitative results.

Overall, we find that
our method
is able to accurately estimate the object pose across most of the frames when no occlusions are present ($98.2\%$ average $\mathrm{ADD(-S)}$
over
the $5$ objects) and
shows
better robustness and accuracy
compared to
the available state-of-the-art CAD-model-free baselines, particularly
when handling
occlusions and objects with limited texture or complex geometry.

%% file: figures/failure_mode_reconstruction_linemod.tex
\begin{figure}
    \captionsetup[subfigure]{labelformat=empty}
    \captionsetup[subfloat]{captionskip=1pt}
    \centering
    \vspace{-5pt}
    \subfloat[{\scriptsize$\mathrm{ape}$: $\SI{1.1}{\milli\meter}$}]{\includegraphics[width=0.32\linewidth]{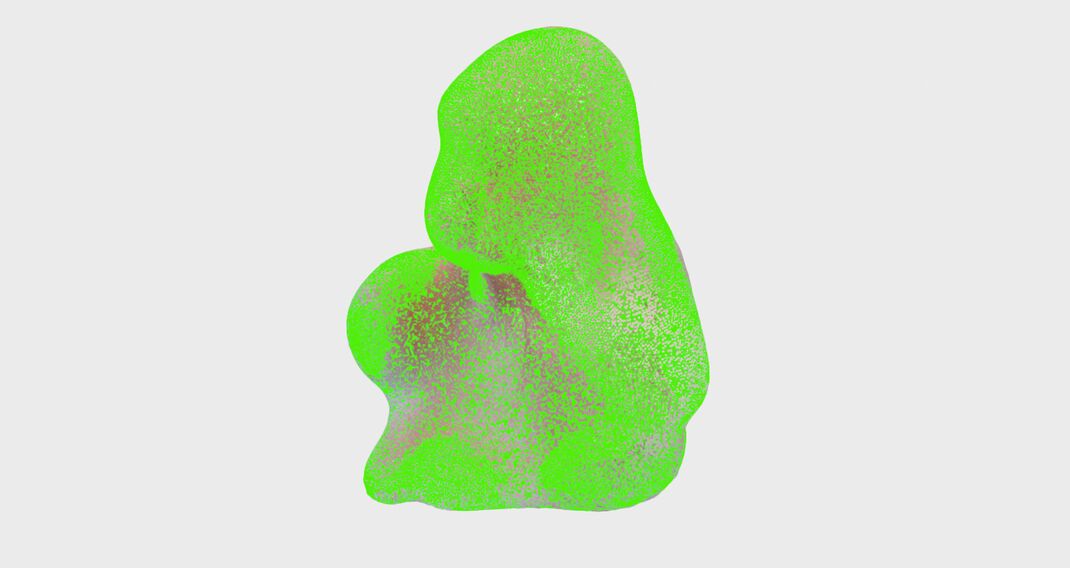}}
    \hspace{0.005\linewidth}
    \subfloat[{\scriptsize$\mathrm{cat}$: $\SI{0.5}{\milli\meter}$}]{\includegraphics[width=0.32\linewidth]{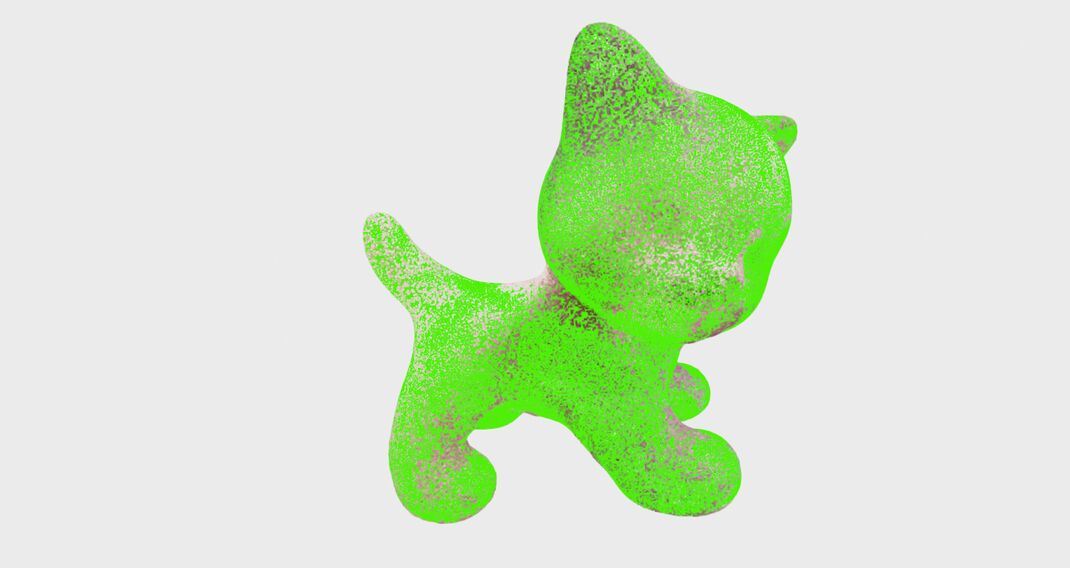}}\\[-5pt]
    \subfloat[{\scriptsize$\mathrm{driller}$: $\SI{0.7}{\milli\meter}$}]{\includegraphics[width=0.32\linewidth]{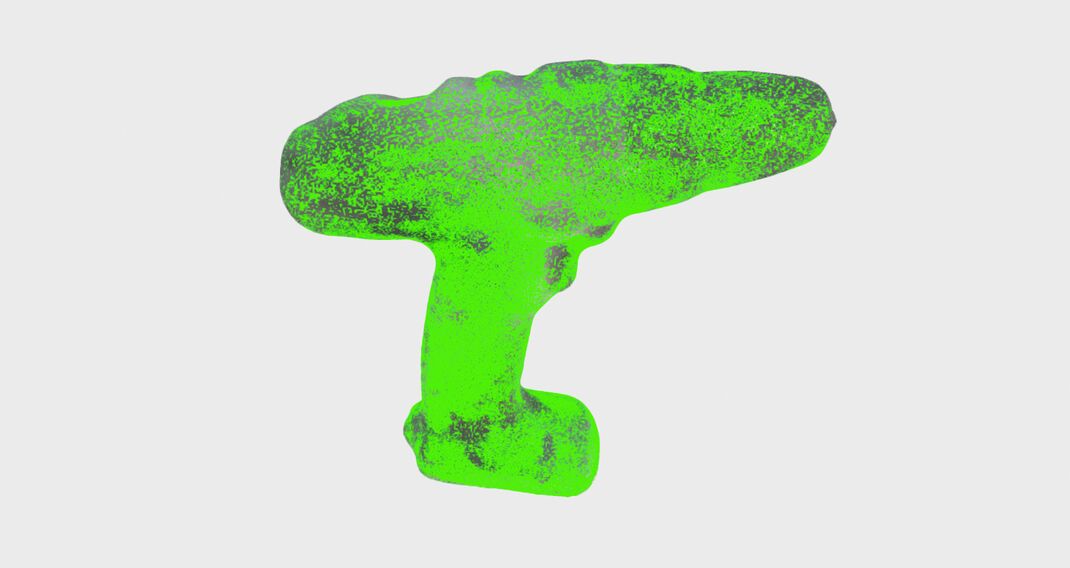}} 
    \hfill
    \subfloat[{\scriptsize$\mathrm{duck}$: $\SI{1.2}{\milli\meter}$}]{\includegraphics[width=0.32\linewidth]{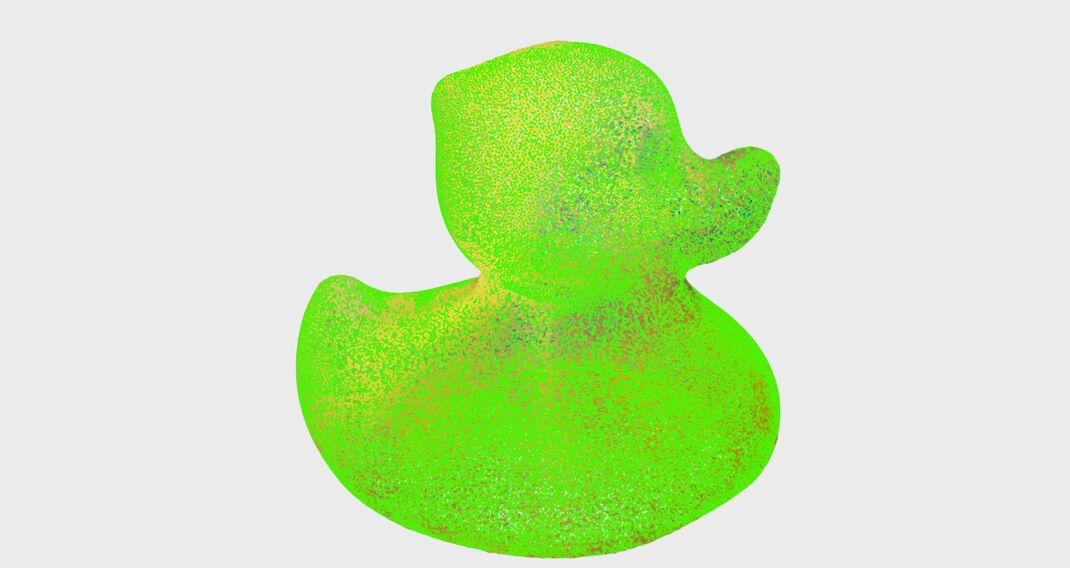}}
    \hfill
    \subfloat[{\scriptsize$\mathrm{glue}$: $\SI{0.5}{\milli\meter}$}]{\includegraphics[width=0.32\linewidth]{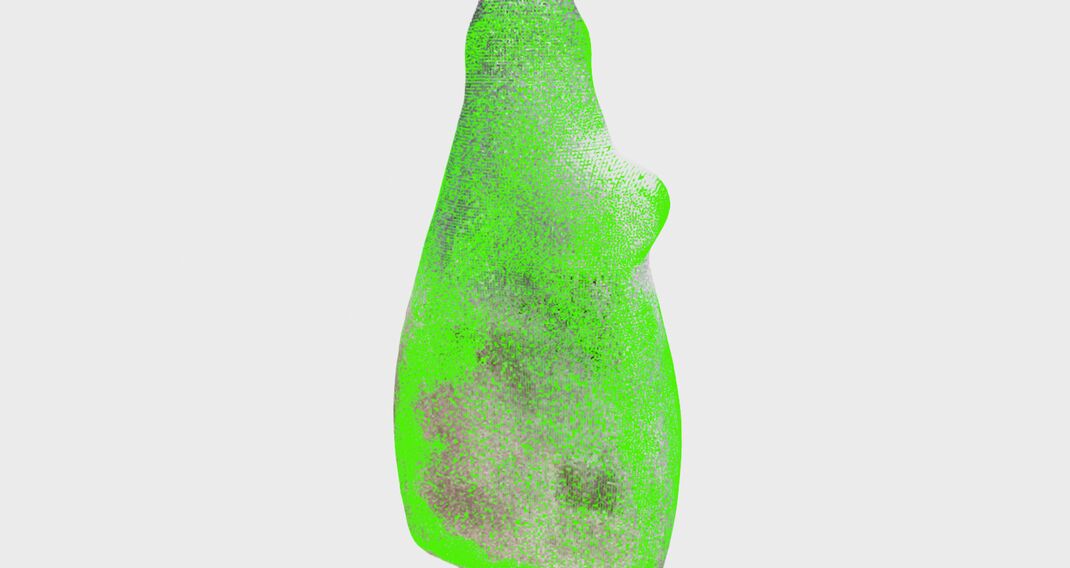}}\\[-5pt]
    \subfloat[{\scriptsize$\mathrm{can}$: $\SI{8.1}{\milli\meter}$}]{\includegraphics[width=0.32\linewidth]{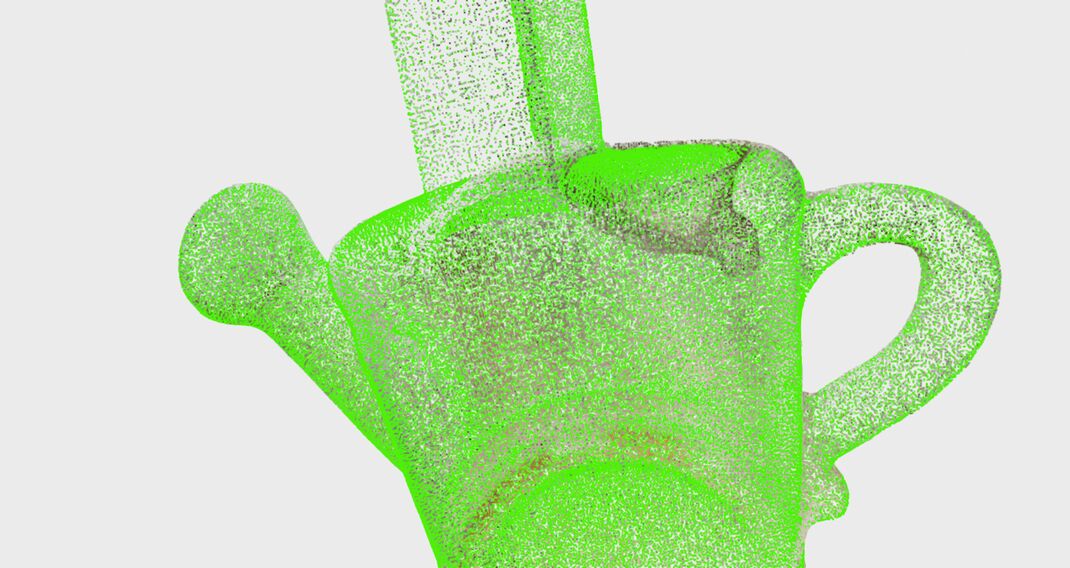}}
    \hfill
    \subfloat[{\scriptsize$\mathrm{eggbox}$: $\SI{7.9}{\milli\meter}$}]{\includegraphics[width=0.32\linewidth]{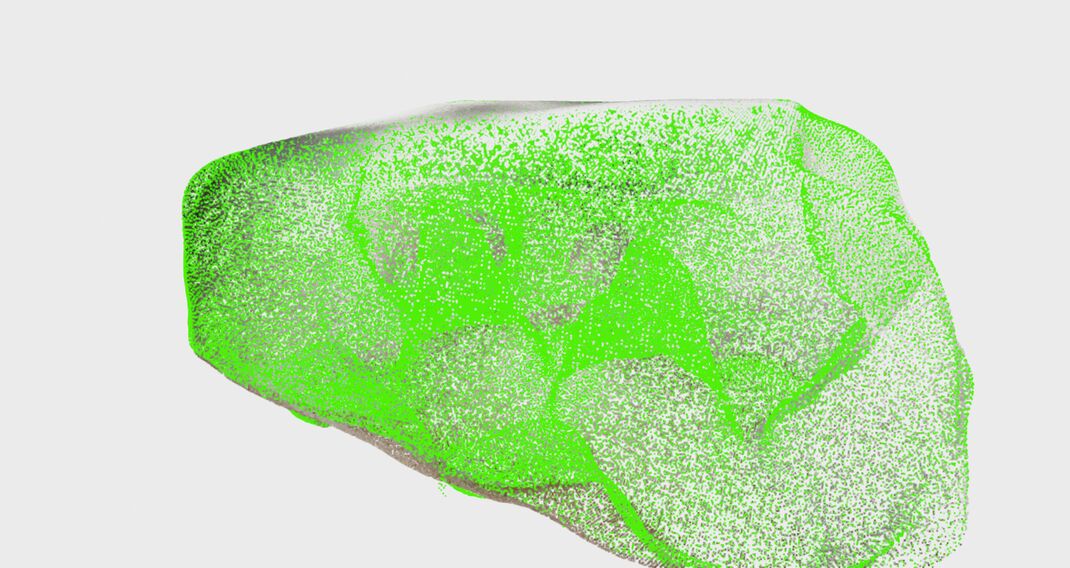}}
    \hfill
    \subfloat[{\scriptsize $\mathrm{holepuncher}$: $\SI{30.3}{\milli\meter}$}]{\includegraphics[width=0.32\linewidth]{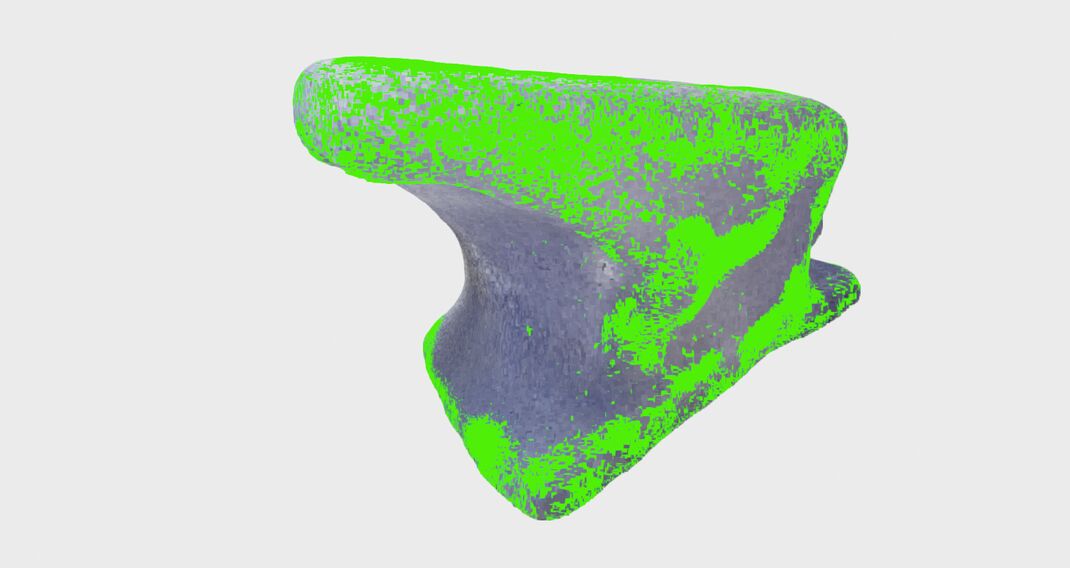}}
    \caption[]{Example NeuS2 reconstructions (shown as textured point cloud), overlaid on the point cloud sampled from the CAD model (shown in green) on the
    objects from LINEMOD-Occlusion.
    Next to the object names we report the forward Chamfer distance with respect to the
    CAD model.
    }
    \label{fig:failure_mode_reconstruction_linemod}
    \vspace{5pt}
\end{figure}

%% file: tables/linemod_occlusion.tex
\begin{table}[t]
\centering
\resizebox{1\linewidth}{!}{

\begin{tabular}{lcccccc}
\toprule
\multirow{2}{*}{Method} & \multirowcell{2}{Training\\renderer} & \multirowcell{2}{Training\\object model} & \multirowcell{2}{Training\\images} & \multicolumn{2}{c}{$\mathrm{AR}_\mathrm{BOP}$}\\
\cmidrule(r){5-6}
& & & & RGB & RGB-D\\
\midrule
\multirow{4}{*}{\approachname} &
NeuS2 & NeuS2 & NeuS2~($10\mathrm{k}$) & $0.554$ & $0.666$ \\
& GL-based & NeuS2 & NeuS2~($10\mathrm{k}$) & $0.570$ & $0.681$ \\
& GL-based & NeuS2 & PBR & $0.646$ & $0.752$ \\
& GL-based & CAD & NeuS2~($10\mathrm{k}$) & $0.568$ & $0.678$ \\
\arrayrulecolor{black!20}\specialrule{0.2pt}{0.2pt}{0.2pt}
\arrayrulecolor{black}
SurfEmb~\cite{Haugaard2022SurfEmb} & GL-based & CAD & PBR & $\mathbf{0.656}$ & $\mathbf{0.758}$\\
EPOS~\cite{Hodan2020EPOS} & - & CAD & PBR & 0.547 & -\\
CDPNv2~\cite{Li2019CDPN} & - & CAD & PBR & 0.624 & - \\
PVNet~\cite{Peng2019PVNet} & - & CAD & PBR & 0.575 & - \\
CosyPose~\cite{Labbe2020CosyPose} & - & CAD & PBR & 0.633 & 0.714 \\
\bottomrule
\end{tabular}
}
\caption{Pose estimation performance averaged across the objects, LINEMOD-Occlusion.%
The baseline results are
taken
from~\cite{Haugaard2022SurfEmb}.}
\label{tab:linemod_occlusion}
\end{table}

%% file: tables/ablation_object_model_for_pose_estimation.tex
\begin{table*}[ht!]
\centering
\resizebox{0.75\linewidth}{!}{
\begin{tabular}{cccccccccccc}
\toprule
\multirow{3}{*}{Training images} & \multicolumn{2}{c}{Object model} & \multicolumn{9}{c}{$\mathrm{AR}_\mathrm{BOP}$}\\
\cmidrule(r){2-3} \cmidrule(r){4-12}
& Training & Pose estimation & $\mathrm{ape}$ & $\mathrm{can}$ & $\mathrm{cat}$ & $\mathrm{driller}$ & $\mathrm{duck}$ & $\mathrm{eggbox}$ & $\mathrm{glue}$ & $\mathrm{holepuncher}$ & $\mathrm{AVG}$ Scene \\
\midrule
NeuS2~($10\mathrm{k}$) & NeuS2 & NeuS2 & 0.519 & 0.761 & 0.467 & 0.604 & 0.638 & 0.284 & 0.546 & 0.695 & 0.570 \\
NeuS2~($10\mathrm{k}$) & NeuS2 & CAD & 0.527 & 0.751 & 0.468 & 0.604 & 0.652 & 0.251 & 0.562 & 0.438 & 0.534 \\
PBR & NeuS2 & NeuS2 & 0.627 & 0.790 & 0.609 & 0.806 & 0.633 & 0.409 & 0.635 & 0.622 & 0.646 \\
PBR & NeuS2 & CAD & 0.613& 0.776 & 0.621 & 0.805 & 0.650 & 0.291 & 0.647 & 0.465 & 0.610 \\
\arrayrulecolor{black!20}\specialrule{0.2pt}{0.2pt}{0.2pt}
\arrayrulecolor{black}
NeuS2~($10\mathrm{k}$) & CAD & CAD & 0.539 & 0.759 & 0.487 & 0.554 & 0.639 & 0.270 & 0.576 & 0.688 & 0.568 \\
NeuS2~($10\mathrm{k}$) & CAD & NeuS2 & 0.534 & 0.709 & 0.486 & 0.549 & 0.619 & 0.290 & 0.554 & 0.622 & 0.549 \\
PBR & CAD & CAD & 0.593 & 0.788 & 0.625 & 0.783 & 0.623 & 0.424 & 0.584 & 0.695 & 0.646 \\
PBR & CAD & NeuS2 & 0.594 & 0.792 & 0.601 & 0.787 & 0.612 & 0.514 & 0.600 & 0.642 & 0.648 \\
\bottomrule
\end{tabular}
}
\caption{Effect of the training images and the object model in pose estimation, LINEMOD-Occlusion.
RGB-only input.
}
\label{tab:ablation_object_model_and_training_images}
\vspace{-15pt}
\end{table*}

%% file: tables/linemod.tex
\begin{table*}[ht!]
\centering
\resizebox{0.75\linewidth}{!}{
\begin{tabular}{ccccccccccc}
\toprule
\multirow{3}{*}{Training object model} & \multirow{3}{*}{Training images} & \multicolumn{9}{c}{$\mathrm{AR}_\mathrm{BOP}$}\\
\cmidrule(r){3-11}
& & $\mathrm{ape}$ & $\mathrm{can}$ & $\mathrm{cat}$ & $\mathrm{driller}$ & $\mathrm{duck}$ & $\mathrm{eggbox}$ & $\mathrm{glue}$ & $\mathrm{holepuncher}$ & $\mathrm{AVG}$ Scenes \\
\midrule
NeuS2 & NeuS2~($10\mathrm{k}$) & 0.782 & 0.957 & 0.863 & 0.930 & 0.790 & 0.880 & 0.781 & 0.842 & 0.853 \\
NeuS2 & PBR & 0.865 & 0.897 & 0.857 & 0.939 & 0.833 & 0.779 & 0.779 & 0.695 & 0.831 \\
CAD & NeuS2~($10\mathrm{k}$) & 0.788 & 0.948 & 0.871 & 0.909 & 0.797 & 0.943 & 0.822 & 0.845 & 0.865 \\
CAD & PBR & 0.804 & 0.901 & 0.875 & 0.954 & 0.828 & 0.780 & 0.705 & 0.780 & 0.829 \\
\bottomrule
\end{tabular}
}
\caption{Pose estimation performance on the LINEMOD dataset, evaluated for the $8$ objects also contained in LINEMOD-Occlusion. All models use the ModernGL renderer and are evaluated with RGB-only inputs.}
\label{tab:linemod}
\vspace{-20pt}
\end{table*}

%% file: figures/neus2_reconstruction_real_world_experiments.tex
\begin{figure*}[ht!]
\centering
\def\colwidth{0.16\textwidth}
\def\extraspaceaddedwidth{0.001\textwidth}
\newcolumntype{M}[1]{>{\centering\arraybackslash}m{#1}}
\addtolength{\tabcolsep}{-4pt}
\centering
    {
\begin{tabular}{M{\colwidth} M{\colwidth} M{\colwidth} M{\colwidth} M{\colwidth}}
\includegraphics[width=\linewidth]{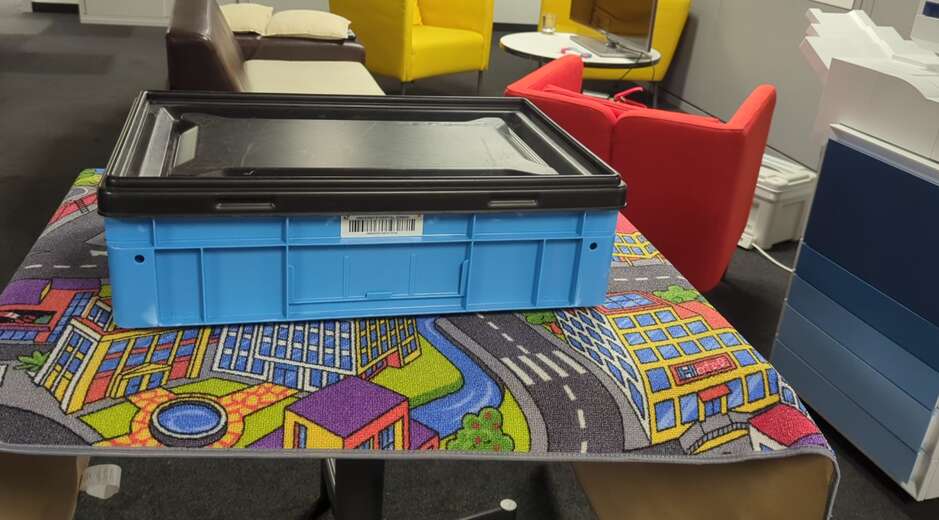} & 
\includegraphics[width=\linewidth]{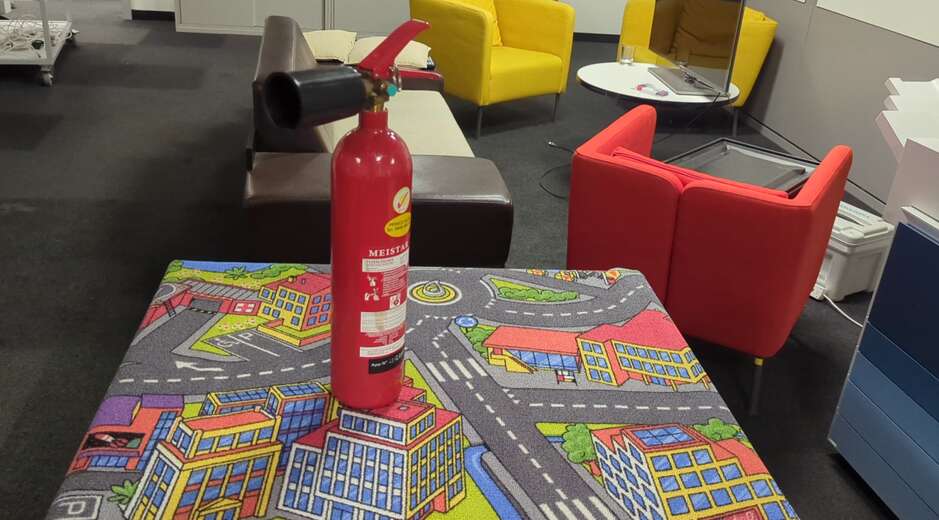} &
\includegraphics[width=\linewidth]{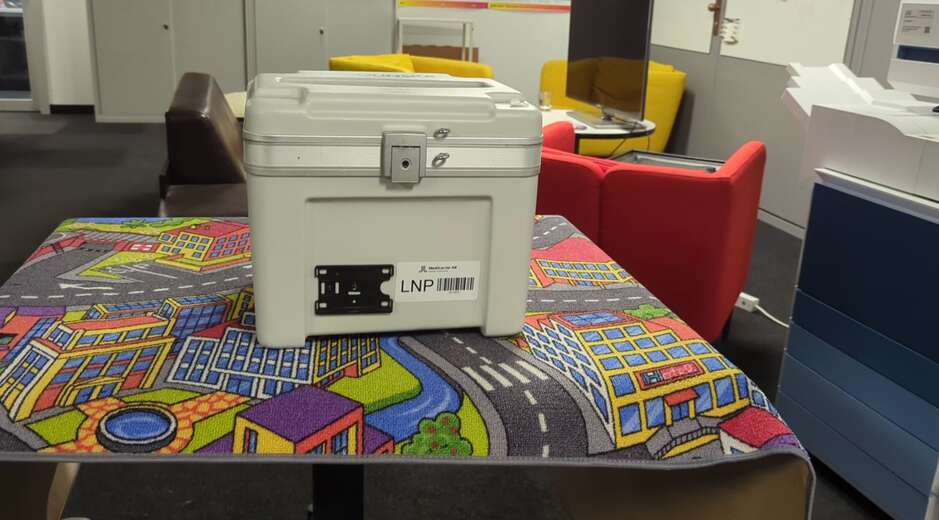} &
\includegraphics[width=\linewidth]{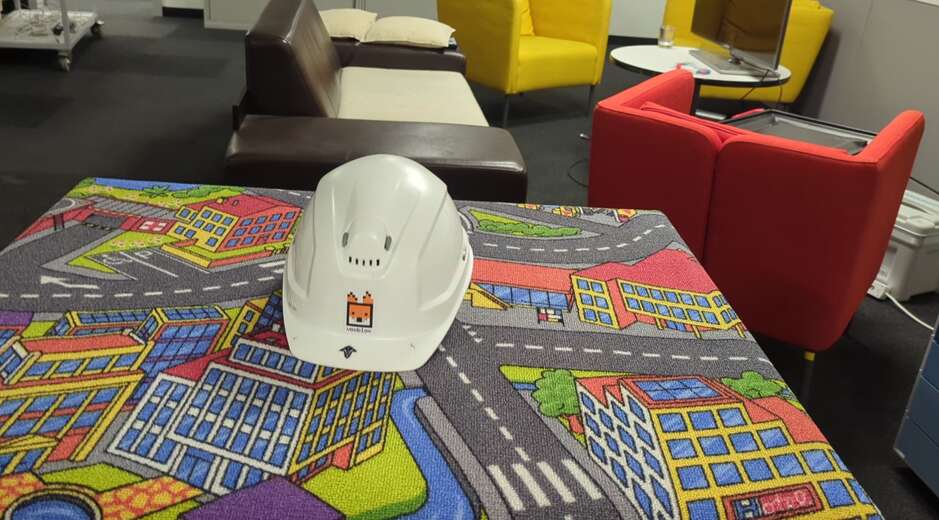} &
\includegraphics[width=\linewidth]{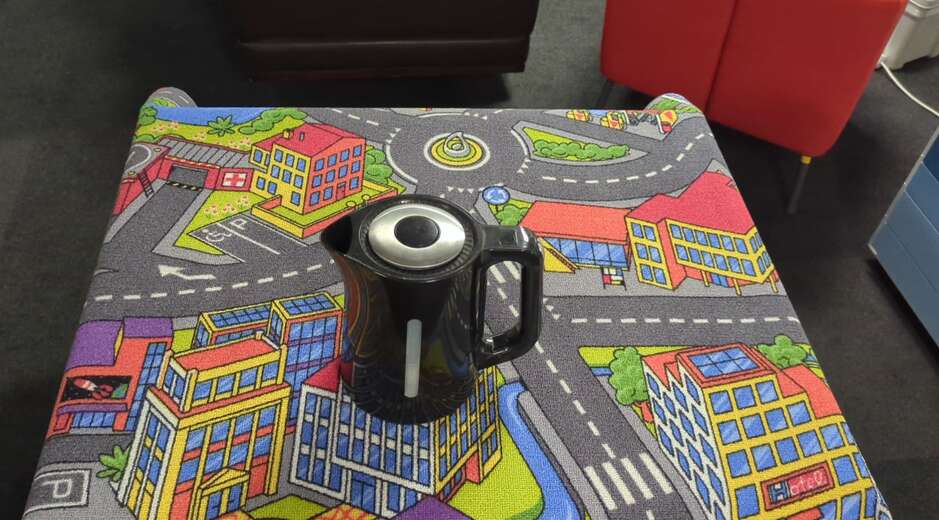}
  \tabularnewline
\includegraphics[width=\linewidth]{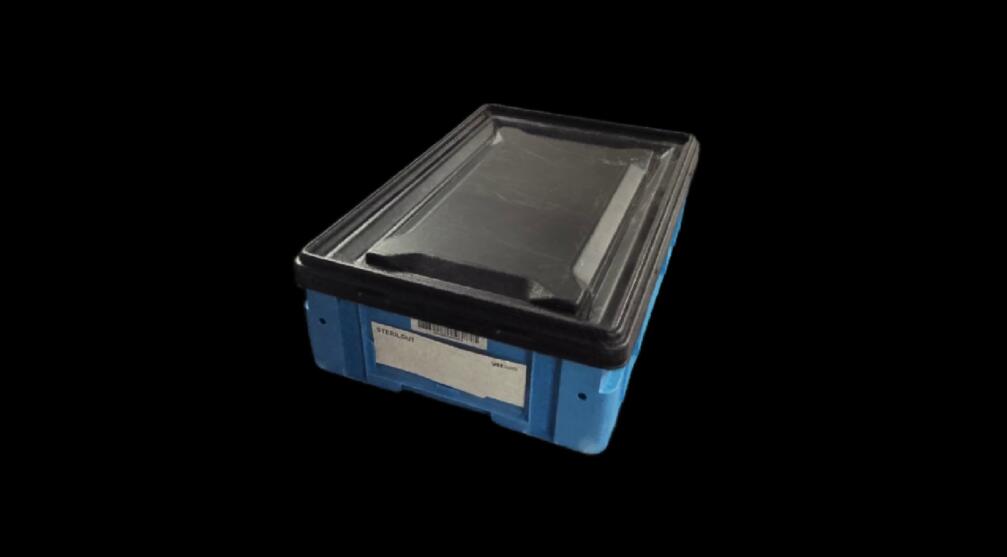} & 
\includegraphics[width=\linewidth]{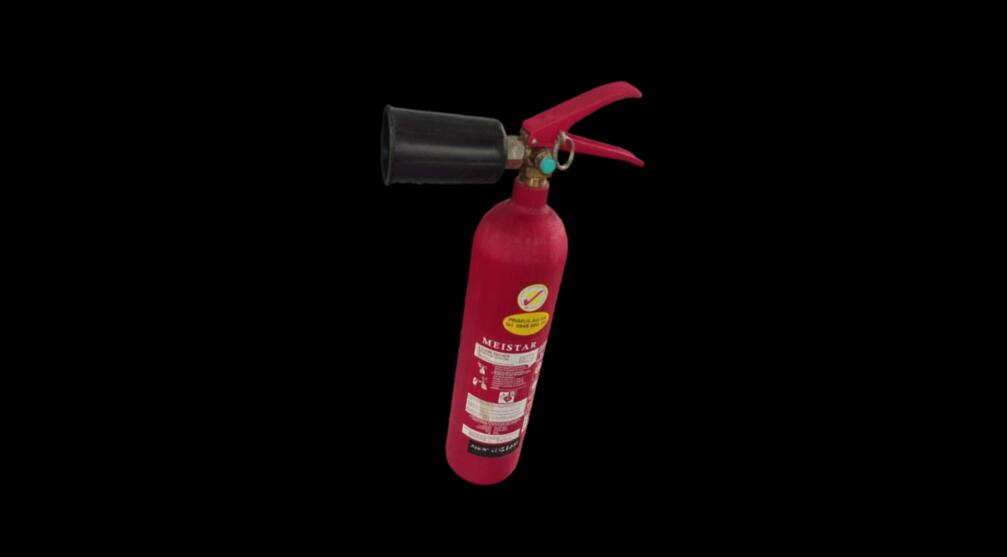} &
\includegraphics[width=\linewidth]{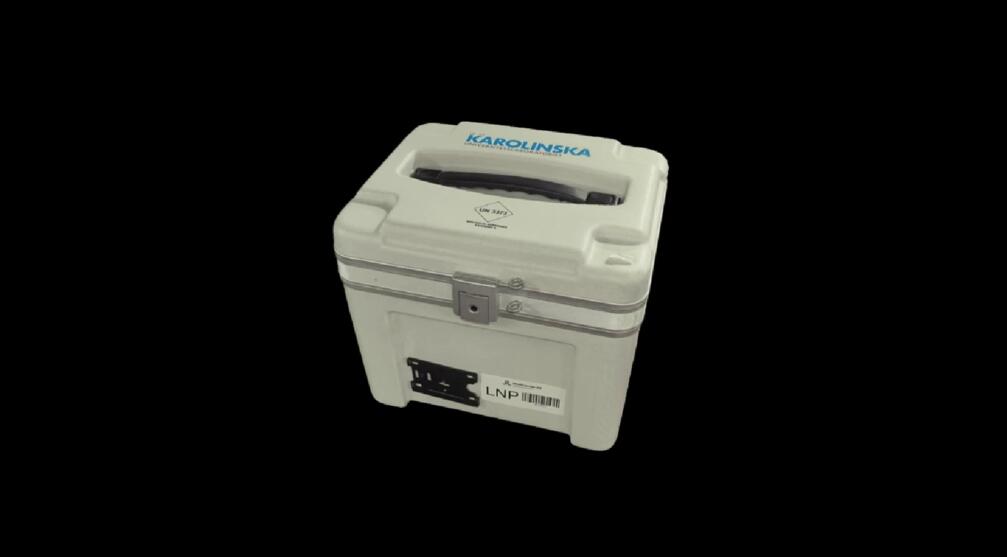} &
\includegraphics[width=\linewidth]{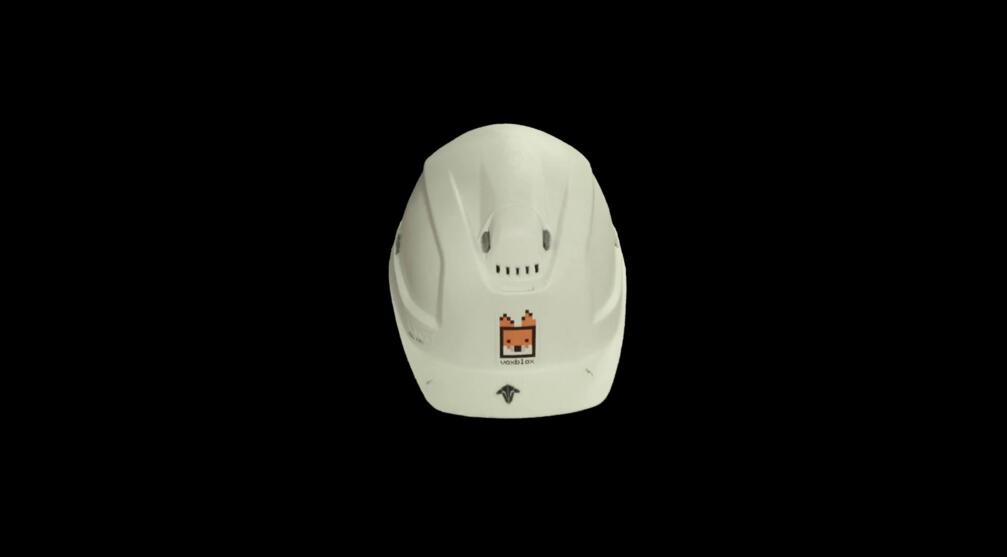} &
\includegraphics[width=\linewidth]{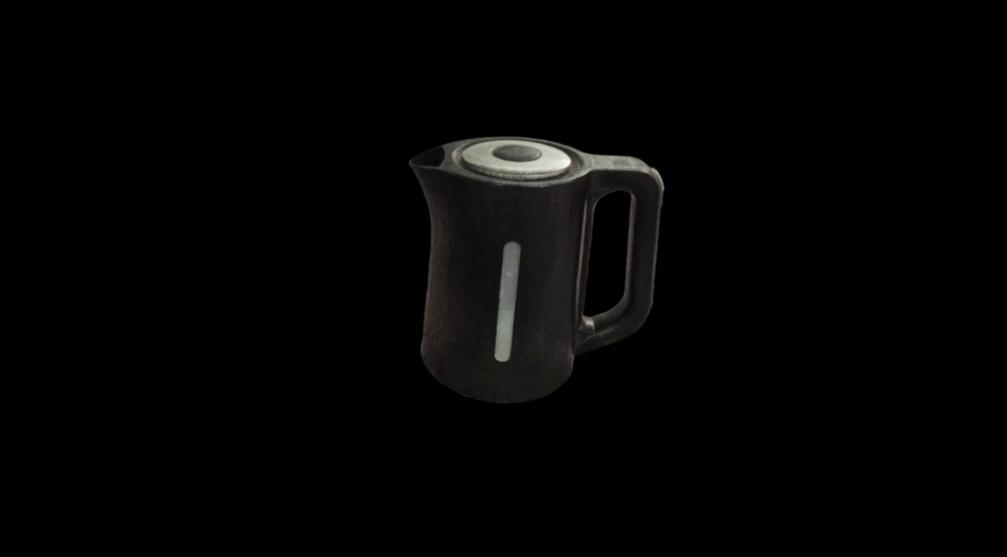}
\tabularnewline
\end{tabular}
}
\caption{Example images captured for model construction in the real-world experiments (top row) and corresponding NeuS2 reconstructions (bottom row). The objects depicted from left to right are: $\mathrm{bluebox}$, $\mathrm{extinguisher}$, $\mathrm{greybox}$, $\mathrm{helmet}$, $\mathrm{kettle}$.
\label{fig:neus2_reconstruction_real_world_experiments}}
\end{figure*}

%% file: figures/qualitative_results.tex
\begin{figure*}[ht!]
\centering
\def\colwidth{0.145\textwidth}
\def\colwidthsem{0.06\textwidth}
\def\extraspaceaddedwidth{0.001\textwidth}
\newcolumntype{M}[1]{>{\centering\arraybackslash}m{#1}}
\addtolength{\tabcolsep}{-4pt}
\centering
    {
\begin{tabular}{M{\colwidthsem} M{\extraspaceaddedwidth} M{\colwidth} M{\colwidth} M{\colwidth} M{\colwidth} M{\colwidth} M{\colwidth}}
\RotText{
{\ssmall \approachname\par (Ours)} 
} & &
\includegraphics[width=\linewidth]{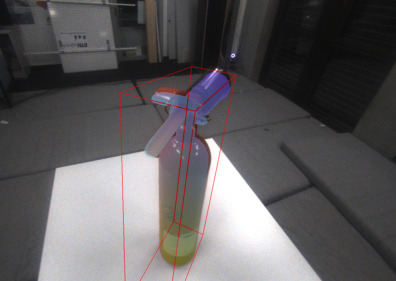} & 
\includegraphics[width=\linewidth]{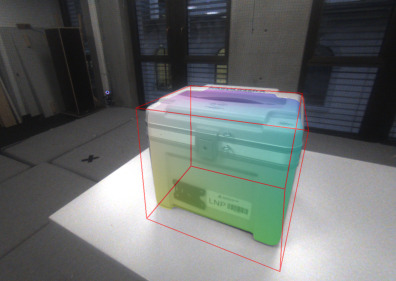} &
\includegraphics[width=\linewidth]{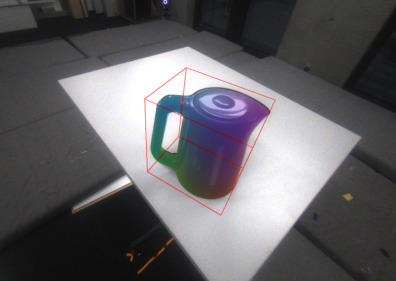} &
\includegraphics[width=\linewidth]{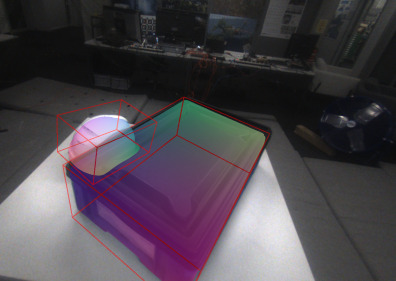} & 
\includegraphics[width=\linewidth]{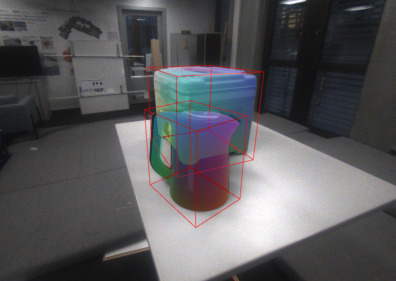} &
\includegraphics[width=\linewidth]{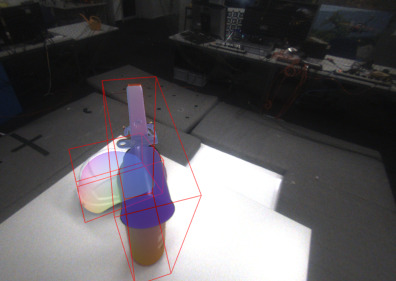}
  \tabularnewline
\RotText{
{\ssmall OnePose++~\cite{He2022OnePose++}\par (w/o tracking,\textrm{ }\textrm{ }\textrm{ }\textrm{ }\textrm{ }\textrm{ }\textrm{ }\textrm{ }\textrm{ }\textrm{ }\textrm{ }\textrm{ }orig. recropping)}
}
& &
\includegraphics[width=\linewidth]{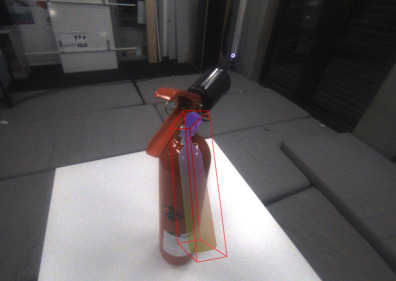} &
\includegraphics[width=\linewidth]{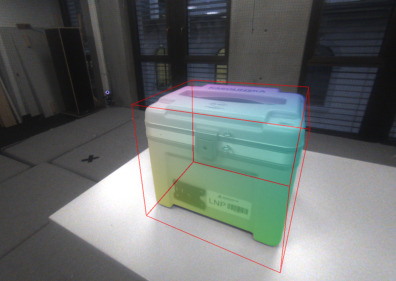} &
\includegraphics[width=\linewidth]{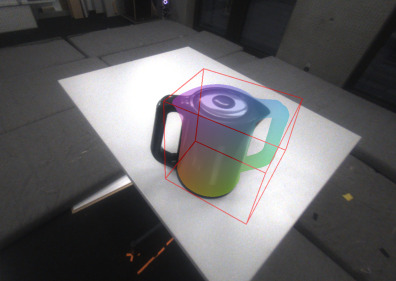} &
\includegraphics[width=\linewidth]{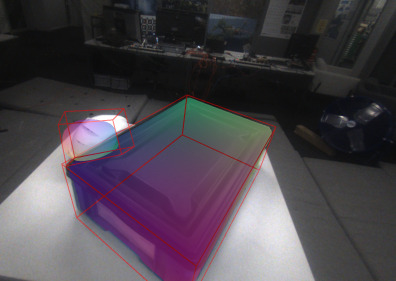} & 
\includegraphics[width=\linewidth]{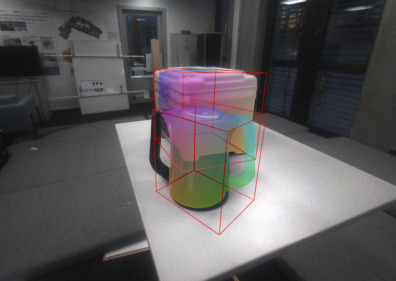} &
\includegraphics[width=\linewidth]{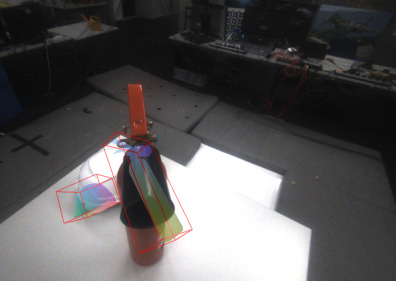}
\tabularnewline
\RotText{
{\ssmall OnePose++~\cite{He2022OnePose++}\par (w/o tracking,\textrm{ }\textrm{ }\textrm{ }\textrm{ }\textrm{ }\textrm{ }\textrm{ }\textrm{ }prop. recropping)}
}
& &
\includegraphics[width=\linewidth]{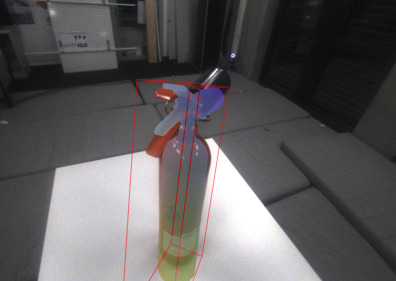} &
\includegraphics[width=\linewidth]{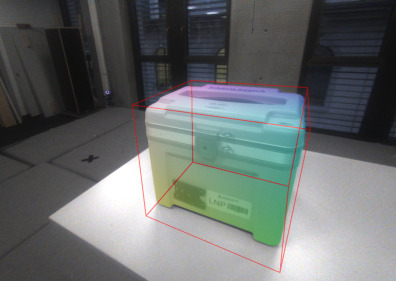} &
\includegraphics[width=\linewidth]{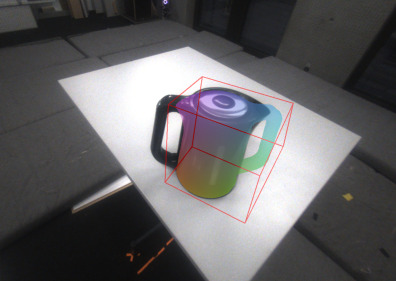} &
\includegraphics[width=\linewidth]{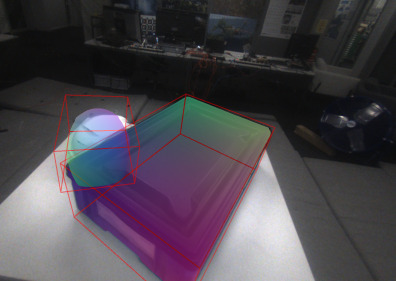} & 
\includegraphics[width=\linewidth]{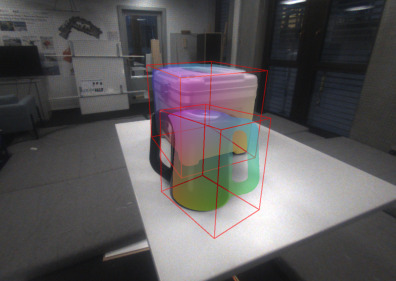} &
\includegraphics[width=\linewidth]{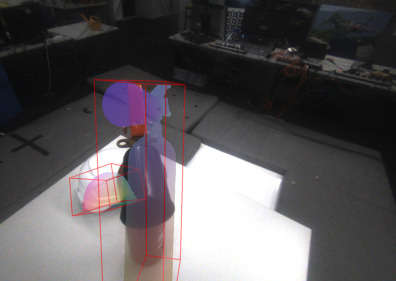}
  \tabularnewline
\RotText{
{\ssmall Gen6D~\cite{Liu2022Gen6D}\par (with tracking)}
}
& &
\includegraphics[width=\linewidth]{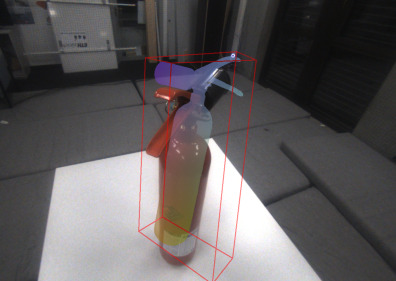} & 
\includegraphics[width=\linewidth]{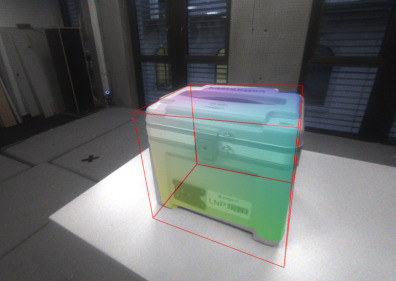} &
\includegraphics[width=\linewidth]{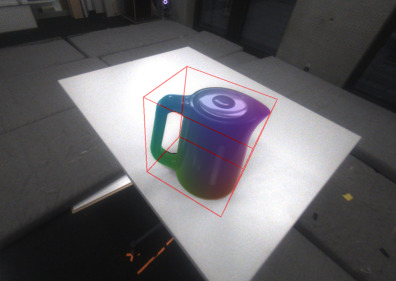} &
\includegraphics[width=\linewidth]{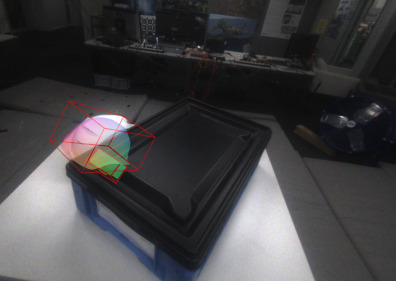} & 
\includegraphics[width=\linewidth]{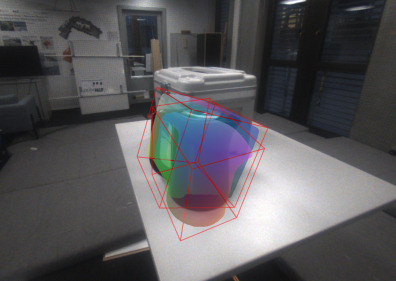} &
\includegraphics[width=\linewidth]{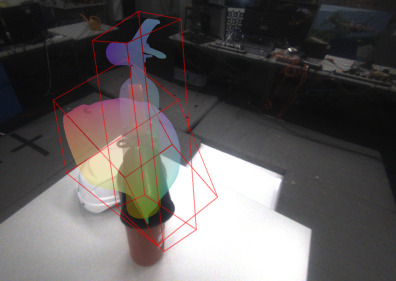}
  \tabularnewline
\RotText{
{\ssmall Gen6D~\cite{Liu2022Gen6D}\par (w/o tracking)}
}
& &
\includegraphics[width=\linewidth]{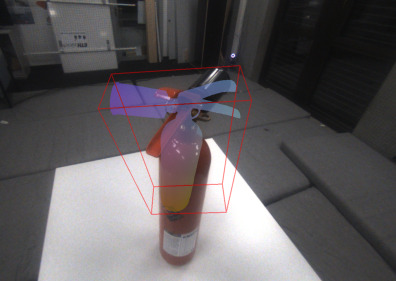} & 
\includegraphics[width=\linewidth]{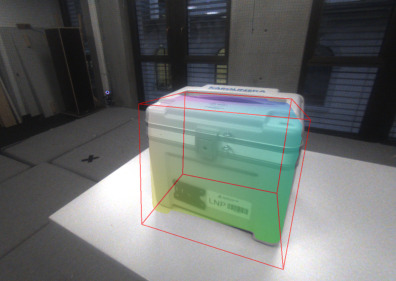} & 
\includegraphics[width=\linewidth]{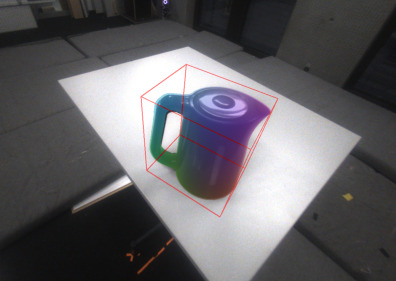} & 
\includegraphics[width=\linewidth]{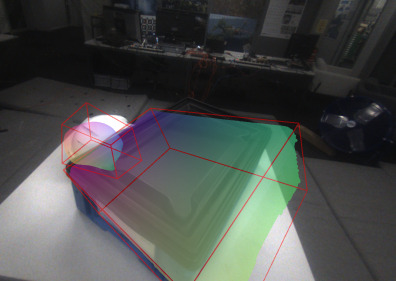} & 
\includegraphics[width=\linewidth]{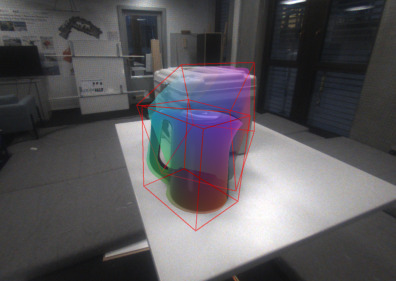} &
\includegraphics[width=\linewidth]{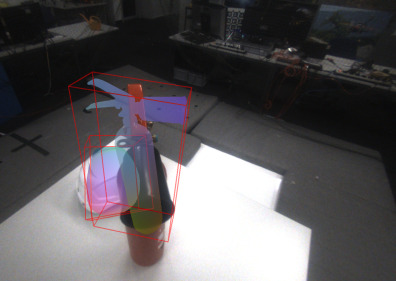}
  \tabularnewline
\end{tabular}
}
\vspace{-4pt}
\caption{Example visualizations of the poses estimated by the different methods in the real-world experiments, displayed as rendered coordinates and reprojected object bounding box overlaid to the original image. The scenes depicted from left to right are: $\mathrm{extinguisher}$, $\mathrm{greybox}$, $\mathrm{kettle}$, \hbox{$\mathrm{bluebox} - \mathrm{helmet}$}, \hbox{$\mathrm{greybox} - \mathrm{kettle}$}, \hbox{$\mathrm{helmet} - \mathrm{extinguisher}$} (cf. Tables~\ref{tab:results_real_world_experiments_no_occlusion} and~\ref{tab:results_real_world_experiments_with_occlusions}).
\label{fig:qualitative_results}}
\vspace{-15pt}
\end{figure*}

%% file: tables/results_real_world_experiments_no_occlusions.tex
\begin{table*}[ht!]
\centering
\resizebox{\linewidth}{!}{
\begin{tabular}{lcccccccccccc}
\toprule
\multirow{3}{*}{Method} & \multicolumn{6}{c}{$\mathrm{ADD(-S)}$} & \multicolumn{6}{c}{$\SI{5}{cm},\ \SI{5}{{}^\circ}$}\\
\cmidrule(lr){2-7} \cmidrule(l){8-13}
& $\mathrm{bluebox}^\star$ & $\mathrm{extinguisher}$ & $\mathrm{greybox}^\star$ & $\mathrm{helmet}$ & $\mathrm{kettle}$ & $\mathrm{AVG}$ Objects & $\mathrm{bluebox}$ & $\mathrm{extinguisher}$ & $\mathrm{greybox}$ & $\mathrm{helmet}$ & $\mathrm{kettle}$ & $\mathrm{AVG}$ Objects \\
\midrule
\approachname (Ours) & \textbf{1.000} & \textbf{0.980} & \textbf{1.000} & \textbf{0.979} & \textbf{0.955} & \textbf{0.982} & \textbf{0.789} & \textbf{0.755} & 0.873 & \textbf{0.937} & \textbf{0.803} & \textbf{0.825} \\
OnePose++~\cite{He2022OnePose++} (original, with tracking) & 0.976 & 0.515 & \textbf{1.000} & 0.629 & 0.371 & 0.683 & 0.476 & 0.388 & 0.853& 0.601& 0.315 & 0.510\\
OnePose++~\cite{He2022OnePose++} (w/o tracking, original recropping) & 0.994 & 0.520 & \textbf{1.000} & 0.629 & 0.315 & 0.676 & 0.530 & 0.388 & \textbf{0.887} & 0.601 & 0.275 & 0.519 \\
OnePose++~\cite{He2022OnePose++} (w/o tracking, proposed recropping) & \textbf{1.000} & 0.791 & \textbf{1.000} & 0.650 & 0.416 & 0.766 & 0.530 & 0.622 & 0.880 & 0.587 & 0.348 & 0.586 \\
Gen6D~\cite{Liu2022Gen6D} (with tracking) & 0.795 & 0.005 & \textbf{1.000} & 0.399 & 0.663 & 0.550 & 0.193 & 0.000 & 0.860 & 0.147 & 0.281 & 0.279 \\
Gen6D~\cite{Liu2022Gen6D} (w/o tracking) & 0.898 & 0.230 & \textbf{1.000} & 0.294 & 0.669 & 0.606 & 0.042 & 0.015 & 0.473 & 0.112 & 0.320 & 0.185 \\
\bottomrule
\end{tabular}
}
\caption{Pose estimation performance on the real-world experiments, non-occluded scenes. ${}^\star$ denotes symmetrical objects.
}
\label{tab:results_real_world_experiments_no_occlusion}
\vspace{-15pt}
\end{table*}

%% file: tables/results_real_world_experiments_with_occlusions.tex
\begin{table*}[ht!]
\centering
\resizebox{\linewidth}{!}{
\begin{tabular}{lcccccccccccccccc}
\toprule
\multirow{4}{*}{Method} & \multicolumn{8}{c}{$\mathrm{AR}_\mathrm{BOP}$} & \multicolumn{8}{c}{$\SI{5}{cm},\ \SI{5}{{}^\circ}$}\\
\cmidrule(r){2-9} \cmidrule{10-17}
& \multicolumn{2}{c}{$\mathrm{bluebox} - \mathrm{helmet}$} & \multicolumn{2}{c}{$\mathrm{greybox} - \mathrm{kettle}$} & \multicolumn{2}{c}{$\mathrm{helmet} - \mathrm{extinguisher}$} & \multicolumn{2}{c}{$\mathrm{kettle} - \mathrm{bluebox}$} & \multicolumn{2}{c}{$\mathrm{bluebox} - \mathrm{helmet}$} & \multicolumn{2}{c}{$\mathrm{greybox} - \mathrm{kettle}$} & \multicolumn{2}{c}{$\mathrm{helmet} - \mathrm{extinguisher}$} & \multicolumn{2}{c}{$\mathrm{kettle} - \mathrm{bluebox}$}\\
\cmidrule(lr){2-3} \cmidrule(r){4-5} \cmidrule(r){6-7} \cmidrule(r){8-9} \cmidrule(lr){10-11} \cmidrule(r){12-13} \cmidrule(r){14-15} \cmidrule(r){16-17}
& $\mathrm{bluebox}^\star$ & $\mathrm{helmet}$ & $\mathrm{greybox}^\star$ & $\mathrm{kettle}$ & $\mathrm{helmet}$ & $\mathrm{extinguisher}$ & $\mathrm{kettle}$ & $\mathrm{bluebox}^\star$ & $\mathrm{bluebox}$ & $\mathrm{helmet}$ & $\mathrm{greybox}$ & $\mathrm{kettle}$ & $\mathrm{helmet}$ & $\mathrm{extinguisher}$ & $\mathrm{kettle}$ & $\mathrm{bluebox}$ \\
\midrule
\approachname (Ours) &  0.937 & \textbf{0.731} & 0.964 & \textbf{0.752} & \textbf{0.918} & \textbf{0.869} & \textbf{0.587} & 0.946 & \textbf{0.632} & \textbf{0.606} & \textbf{0.857} & \textbf{0.528} & \textbf{0.798} & 0.664 & \textbf{0.347} & \textbf{0.748} \\
OnePose++~\cite{He2022OnePose++} (original, with tracking) & 0.985 & 0.496 & 0.982 & 0.155 & 0.522 & 0.587 & 0.271 & 0.934 & 0.588 & 0.394 & 0.833 & 0.056 & 0.479 & 0.521 & 0.068 & 0.401 \\
OnePose++~\cite{He2022OnePose++} (w/o tracking, original recropping) & 0.983 & 0.516 & 0.994 & 0.216 & 0.557 & 0.524 & 0.343 & \textbf{0.991} & \textbf{0.632} & 0.385 & \textbf{0.857} & 0.153 & 0.521 & 0.345 & 0.102 & 0.435 \\
OnePose++~\cite{He2022OnePose++} (w/o tracking, proposed recropping) & \textbf{0.986} & 0.444 & \textbf{0.996} & 0.303 & 0.517 & 0.776 & 0.416 & 0.977 & \textbf{0.623} & 0.269 & 0.833 & 0.194 & 0.445 & \textbf{0.697} & 0.184 & 0.401\\
Gen6D~\cite{Liu2022Gen6D} (with tracking) & 0.002 & 0.393 & 0.753 & 0.237 & 0.150 & 0.124 & 0.056 & 0.151 & 0.000 & 0.087 & 0.476 & 0.056 & 0.059 & 0.000 & 0.000 & 0.000 \\
Gen6D~\cite{Liu2022Gen6D} (w/o tracking) & 0.558 & 0.319& 0.749 & 0.548 & 0.439 & 0.184 & 0.473 & 0.489 & 0.044 & 0.058 & 0.524 & 0.139 & 0.084 & 0.017 & 0.116 & 0.122 \\
\bottomrule
\end{tabular}
}
\caption{Pose estimation performance on the real-world experiments, occluded scenes. ${}^\star$ denotes symmetrical objects.
}
\label{tab:results_real_world_experiments_with_occlusions}
\vspace{-20pt}
\end{table*}

%% file: sections/6_discussion_and_conclusion.tex
\section{Conclusions~\label{sec:conclusion}}
We presented
a pipeline
to train a state-of-the-art
6D
object
pose
estimator
from just a
small set of
input images.
We propose
forming
a NeuS2-based
object
representation
through
semi-automated labeling 
and
generating
photorealistic
training
images
through
NeuS2
rendering
with
simple {cut-and-paste}
augmentation.
These two components obviate the need for both a CAD model and PBR-based image generation, providing a straightforward and practical solution for real-world robotic scenarios.
Our method
shows
competitive performance
with respect to state-of-the-art
approaches
based on
CAD models and PBR synthetic
data.
Finally, our approach
outperforms
the leading
CAD-model-free approaches,
demonstrating
high
accuracy, robustness to mild occlusions,
and
ease of use in
the real world.